\documentclass{article}
\usepackage[left=1in,right=1in,top=1.in,bottom=1.in]{geometry}
\usepackage{color}
\usepackage[utf8]{inputenc}
\usepackage{parskip}

\usepackage{amssymb}
\usepackage{amsbsy}
\usepackage{stmaryrd}
\usepackage{amsmath}
\usepackage{booktabs}
\usepackage{hyperref}
\usepackage{subcaption}

\newenvironment{Proof}{\paragraph{Proof:}}{\hfill$\square$}

\newcommand{\B}{\boldsymbol}

\DeclareMathOperator*{\argmax}{argmax}

\DeclareMathOperator*{\tolim}{\to}

\newtheorem{theorem}{Theorem}
\newtheorem{defn}{Definition}

\newtheorem{asu}{Assumption}

\usepackage{tikz}
\usetikzlibrary{arrows,automata}

\title{Learning higher-order sequential structure with cloned HMMs}
\author{Antoine Dedieu\thanks{Both authors contributed equally to this work.} \thanks{Email: \texttt{\{antoine,nishad\}@vicarious.com}}  , Nishad Gothoskar\footnotemark[1] \footnotemark[2]  , Scott Swingle, Wolfgang Lehrach, \\ Miguel Lázaro-Gredilla, Dileep George}
\date{\today}

\begin{document}

\maketitle

\begin{abstract}
Variable order sequence modeling is an important problem in artificial and natural intelligence. While overcomplete Hidden Markov Models (HMMs), in theory, have the capacity to represent long-term temporal structure, they often fail to learn and converge to local minima. We show that by constraining HMMs with a simple sparsity structure inspired by biology, we can make it learn variable order sequences efficiently. We call this model cloned HMM (CHMM) because the sparsity structure enforces that many hidden states map deterministically to the same emission state. CHMMs with over 1 billion parameters can be efficiently trained on GPUs without being severely affected by the credit diffusion problem of standard HMMs. Unlike n-grams and sequence memoizers, CHMMs can model temporal dependencies at arbitrarily long distances and recognize contexts with ``holes'' in them. Compared to Recurrent Neural Networks and their Long Short-Term Memory extensions (LSTMs), CHMMs are generative models that can natively deal with uncertainty. Moreover, CHMMs return a higher-order graph that represents the temporal structure of the data which can be useful for community detection, and for building hierarchical models. Our experiments show that CHMMs can beat n-grams, sequence memoizers, and LSTMs on character-level language modeling tasks. CHMMs can be a viable alternative to these methods in some tasks that require variable order sequence modeling and the handling of uncertainty.
\end{abstract}

\section{Introduction}
\label{sec:introduction}
Sequence modeling is a fundamental real-world problem with a wide range of applications.
Recurrent neural networks (RNNs) are currently preferred in sequence prediction tasks due to their ability to model long-term and variable order dependencies. However, RNNs have disadvantages in several applications because of their inability to natively handle uncertainty, and because of the inscrutable internal representations.

Probabilistic sequence models like Hidden Markov Models (HMM) have the advantage of more interpretable representations and the ability to handle uncertainty. Although overcomplete HMMs with many more hidden states compared to the observed states can, in theory, model long-term temporal dependencies \cite{overcomplete-hmm-nips},  training HMMs is challenging due to credit diffusion \cite{bengio1995diffusion}. For this reason, simpler and inflexible n-gram models are preferred to HMMs for tasks like language modeling. Tensor decomposition methods \cite{anandkumar2012method} have been suggested for the learning of HMMs in order to overcome the credit diffusion problem, but current methods are not applicable to the overcomplete setting where the full-rank requirements on the transition and emission matrices are not fulfilled. Recently there has been renewed interest in the topic of training overcomplete HMMs for higher-order dependencies with the expectation that sparsity structures could potentially alleviate the credit diffusion problem \cite{overcomplete-hmm-nips}.

\begin{figure}[t!]
    \centering
    \includegraphics[width=5.25in]{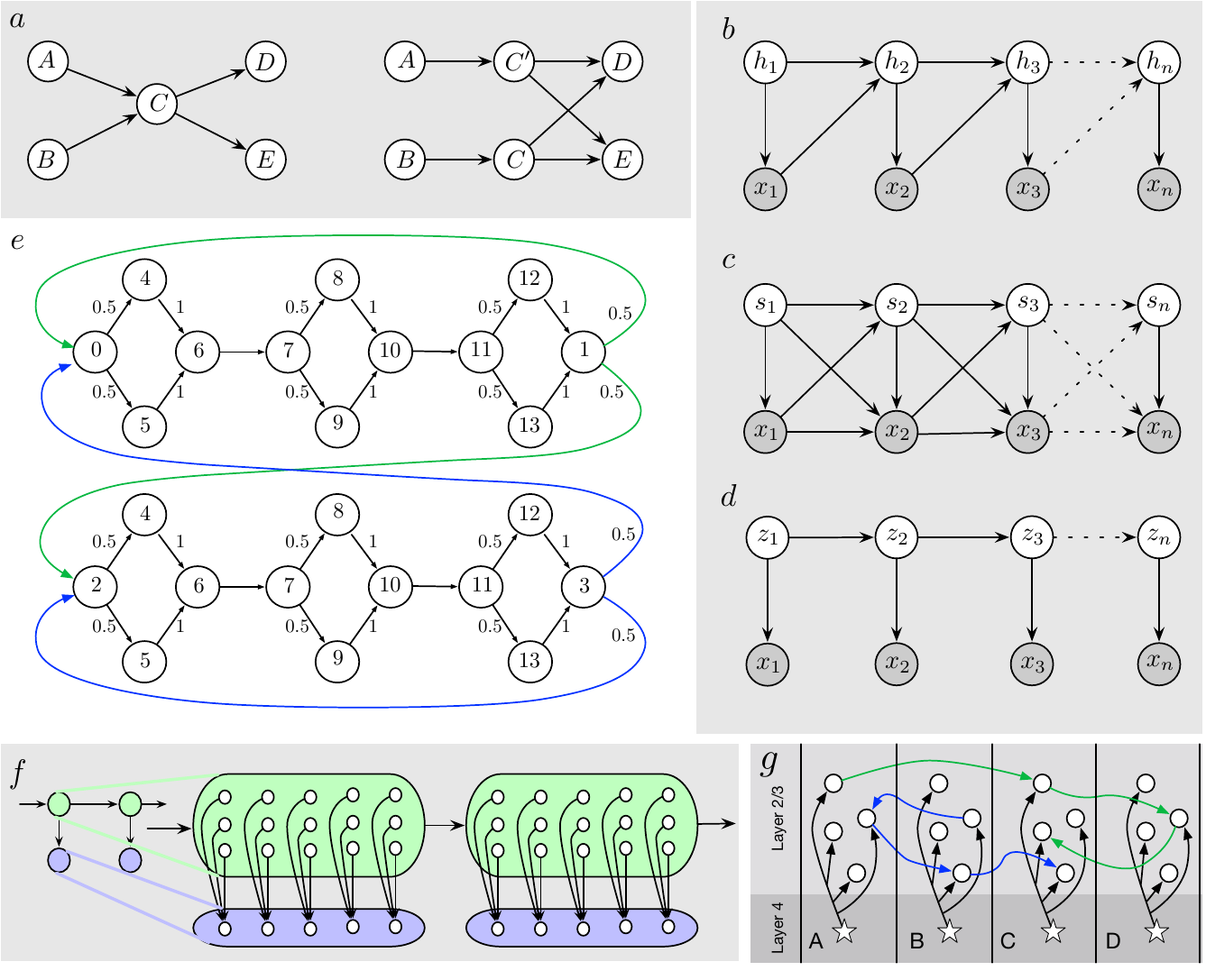}
    \caption{Structure of a CHMM. (a) A first order Markov model (left) constructed from sequences A-C-E and B-C-D cannot capture the higher-order dependencies. Cloning the state C (right) helps to capture this higher order information \cite{cormack1987data}. (b)  Graphical model for a FSM.  (c, d) Graphical models corresponding to a CHMM. (See text for details). (e) The finite state machine used in the toy example (f) Visualization of the emission structure of a CHMM. (g) The columnar organization of the neocortex is postulated as reflecting the cloning structure. Pyramidal cells in layer 2/3 of the cortex receive within-column vertical projections from layer 4. The lateral connections that cross column boundaries, but stay within the laminae encode the sequences \cite{hawkins2009sequence, george2009towards}. This figure shows sequences A-C-D-C (green arrows) and B-A-B-C (blue arrows). Different copies of a cell are used to represent the different sequential contexts.  }
    \label{fig:1}
\end{figure}

In this paper we demonstrate that a particular sparsity structure on the emission matrix can help HMMs learn higher-order temporal structure using the standard Expectation-Maximization algorithms \cite{EM-cv} (Baum-Welch) and its online variants. We call this model cloned HMM (CHMM) because the sparsity structure deterministically maps multiple hidden states (clones) to the same emission state, whereas the emission matrix in a standard HMM is dense and allows for any hidden state to emit any emission state. The basic idea originated in a popular compression method called dynamic Markov coding (DMC) \cite{cormack1987data}, where the temporal dependence in a first order Markov chain is gradually increased by `cloning' the states Fig \ref{fig:1}(a). The same idea has been re-discovered several times in different domains \cite{hawkins2009sequence, xu2016representing,cui2016continuous, persson2016maps}. Cloned hidden states are conjectured to be behind the higher-order sequence representations in bird songs \cite{okubo2015growth}, and specific aspects of the laminar and columnar organization of cortical circuits are postulated to reflect this cloning structure \cite{hawkins2009sequence, george2009towards} (Fig \ref{fig:1}(g)). Researchers have attempted to learn such structures, or partially observable Markov models, using greedy state splitting algorithms that start from a first-order Markov model \cite{cormack1987data, callut2005inducing, persson2016maps, xu2016representing}. Instead, we recognize that the cloning structure can be initialized as a sparse emission matrix of the HMM. The HMM allocates a maximum split capacity for each symbol up front and lets the learning algorithm decide how to utilize that capacity. This results in more flexibility compared to greedy state-splitting approaches (see Appendix \ref{sec:greedysplitting}).

We test the efficacy of CHMMs by using them to learn character-level models of English language using a variety of English text datasets. Our experiments show that CHMMs outperform standard HMMs, n-grams, sequence memoizers, and LSTM-RNNs on several of these tasks while being efficient to train and infer.  The models learned by CHMM are overcomplete by a factor of $~1000$, are extremely sparse, and can be thought of as learning the underlying structure of sequence generation. Unlike n-grams, CHMMs can model arbitrarily long temporal dependencies without having to specify the order up front. When a predictive context contains ``holes'' of irrelevant symbols\footnote{The term ``holes'' has been used \cite{langhabel2017feature} to refer to parts of a context that are irrelevant for prediction. For instance, we can consider a data set in which the context \emph{I am going} will predict \emph{to} regardless of the existence of additional intermediate symbols, so that other contexts like \emph{I kgt am mnbh going} or \emph{I mqlr am pzb going} should result in the same predictive distribution. We would refer to the groups of inserted symbols \emph{kgt}, \emph{mnbh} as ``holes''.} that do not affect prediction, n-grams and sequence memoizers \cite{sequence-memoizer} will need exponentially (in the number of holes) more training data to be able to use the context for prediction, whereas CHMMs can handle such sequences efficiently and flexibly. 
Being generative models, they can model uncertainty natively, and can answer queries that were not part of the training objective. The learned transition matrix of a CHMM is a sparse graph of the generation process, and structural analysis of this graph may reveal communities that can help with hierarchical modeling and planning.

The rest of this paper is organized as follows: In Section \ref{sec:example} we introduce a toy example to illustrate the representational properties of CHMM, and discuss related work. In Section \ref{sec:model} we define the CHMM model and briefly discuss the variations of the EM algorithm we use to learn its parameters. Our results (Section \ref{sec:experiments}) are divided into two subsections. In the first subsection we use synthetic data with known optimal solutions to demonstrate the properties of CHMMs in comparison with regular HMMs and LSTM-RNNs. In the second subsection we use a variety of English text datasets to compare CHMMs with n-grams, sequence memoizers, and LSTM-RNNs. We conclude in Section \ref{sec:discussion} with a discussion of future work in the context of our findings.

\section{Modeling a toy sequence}
\label{sec:example}
Consider the finite state machine\footnote{FSMs can be classified in nondeterministic finite automata (NFA) and deterministic finite automata (DFA). While it is possible to convert between these two representations, the first one is exponentially more compact for some FSMs. The formulation presented here applies to both categories, thus allowing the most compact representation to be used.} (FSM) depicted in Fig. \ref{fig:1}(e). Each of the 24 nodes corresponds to a different internal state. Arrows are labeled with the probability of transition between two given states. The numbers inside the nodes indicate the symbol that the machine accepts when transitioning out of that node. E.g., when at a node labeled with a 7 (note that there are two internal states with this label), the machine will only accept a 7 as the input, and it will transition with equal probability to a state that accepts either an 8 or a 9. The output sequence of this FSM has interesting long-term structure. For example, observing a 0 guarantees a 1 appearing eight time-steps later. Similarly, a 2 will predict a 3 eight steps later with absolute certainty, despite the stochastic transitions in between. The task at hand is to discover this machine (or an equivalent one) based on a long-enough sequence generated by it.

We could model this sequence as a generic FSM, with the corresponding Bayesian network (BN) shown in Fig. \ref{fig:1}(b), where $h_n$ is the hidden state of the machine at time $n$, and $x_n$ is the observation when transitioning out of $h_n$ towards $h_{n+1}$. In particular, it can be modelled when the dimension of the hidden space $h_n$ (number of different discrete values that it can take) is of size 24, which is equal to the number of nodes \footnote{We can exploit some properties of the sequence considered herein to slightly restrict the hidden space size.}. We would then need to learn the conditional probabilities that each arrow represents. We only need to learn one time slice, since they are all identical.

Another way to encode the FSM as a BN is depicted in Fig. \ref{fig:1}{c}.
The additional arrows with respect to the FSM model increase its expressiveness and now we can encode our toy sequence with \emph{only} 2 hidden states, instead of 24. Since now we can also refer to the previous emission and state to decide the current emission, the hidden state no longer needs to remember in which of the 24 hidden states it is, but only in which of the 2 copies (the top or bottom row of Fig. \ref{fig:1}(e) that emit a given symbol) it is. We refer to each hidden state that $s_n$ can take as a different \emph{clone} of the emission $x_n$.

It is possible to perform exact inference on this model, as well as to learn the conditional probabilities that define it using expectation maximization (EM). To illustrate this, note that we can collect $z_n \equiv (s_n, x_n)$ and write the model as in Fig. \ref{fig:1}(d). This graphical model is a hidden Markov model \emph{with a fixed emission matrix}, since given how we created $z_n$, the emissions follow deterministically from the state. Now each group of hidden states that result in the same emission is a group of \emph{clones}, since they look the same when emitted. For this reason, we call it the cloned HMM (CHMM). The number of clones associated to each emission is fixed, but can be different per emission. The transition matrix encodes the conditionals. This means that standard Baum-Welch (minus the learning of the emission matrix) can be used to recover the above FSM when using enough clones (in theory only two clones are required, but in practice more clones can help escape local minima).

To be completely clear, Fig \ref{fig:1}(c) and  Fig \ref{fig:1}(d) are two representations of the same model. The semantics of the hidden variable differs, however: in this example, the hidden states $s$ in Fig \ref{fig:1}(c) have cardinality 2, whereas the hidden states $z$  Fig  \ref{fig:1}(d) have cardinality 24, since the latter collect both $s$ and $x$.

Fig \ref{fig:1}(f) illustrates how the different states in an HMM are tied to the same emission state. On the left is the graphical model for an HMM. On the right we show the details of the internals within each node. This CHMM is overcomplete with three times as many hidden states (shown as white circles within the green oval) as emission states (shown as white circles within the blue oval). The emission matrix, comprising the arrows from the hidden states to the emission states, shows the cloned structure with multiple hidden states mapping to the same emission state. This matrix is initialized deterministically and does not change during learning. The transition matrix, encapsulated in the arrow between the green ovals, is initialized randomly and is modified during learning. The set of hidden states that deterministically emit the same observation can be considered as ``clones'' of each other and of the observed symbol. The different clones, through the learning of the transition matrix, will come to represent the different temporal contexts in which a particular symbol occurs in the training data. The long range dependencies get embedded in the hidden state and are propagated at arbitrary distances, thus becoming Markovian in that hidden state. 

\begin{figure}
\begin{subfigure}{0.5\textwidth}
    \centering
    \includegraphics[width=3in]{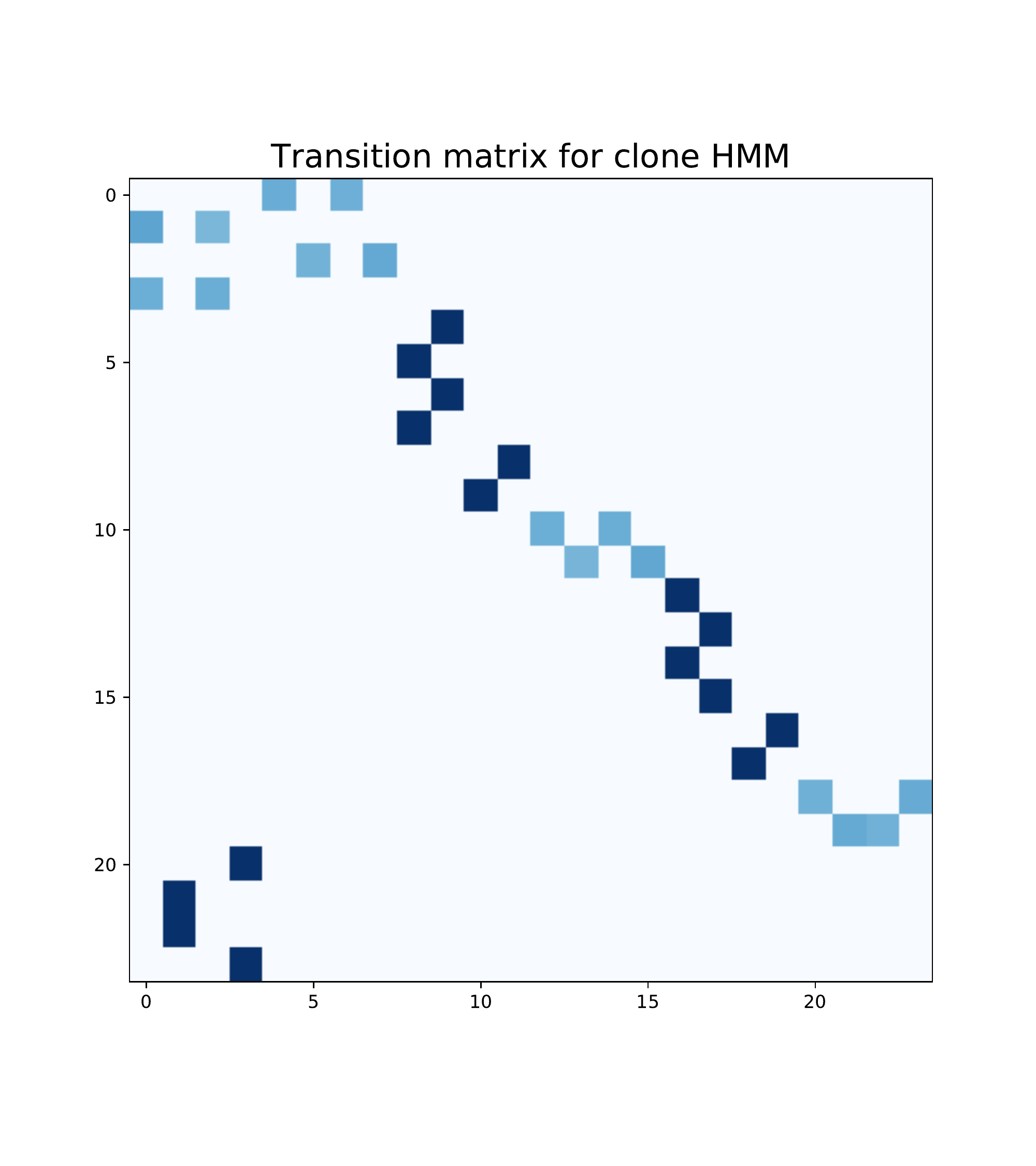}
        \vspace{-1cm}

    \caption{CHMM}
    
\end{subfigure}
\begin{subfigure}{0.5\textwidth}
    \centering
    \includegraphics[width=3in]{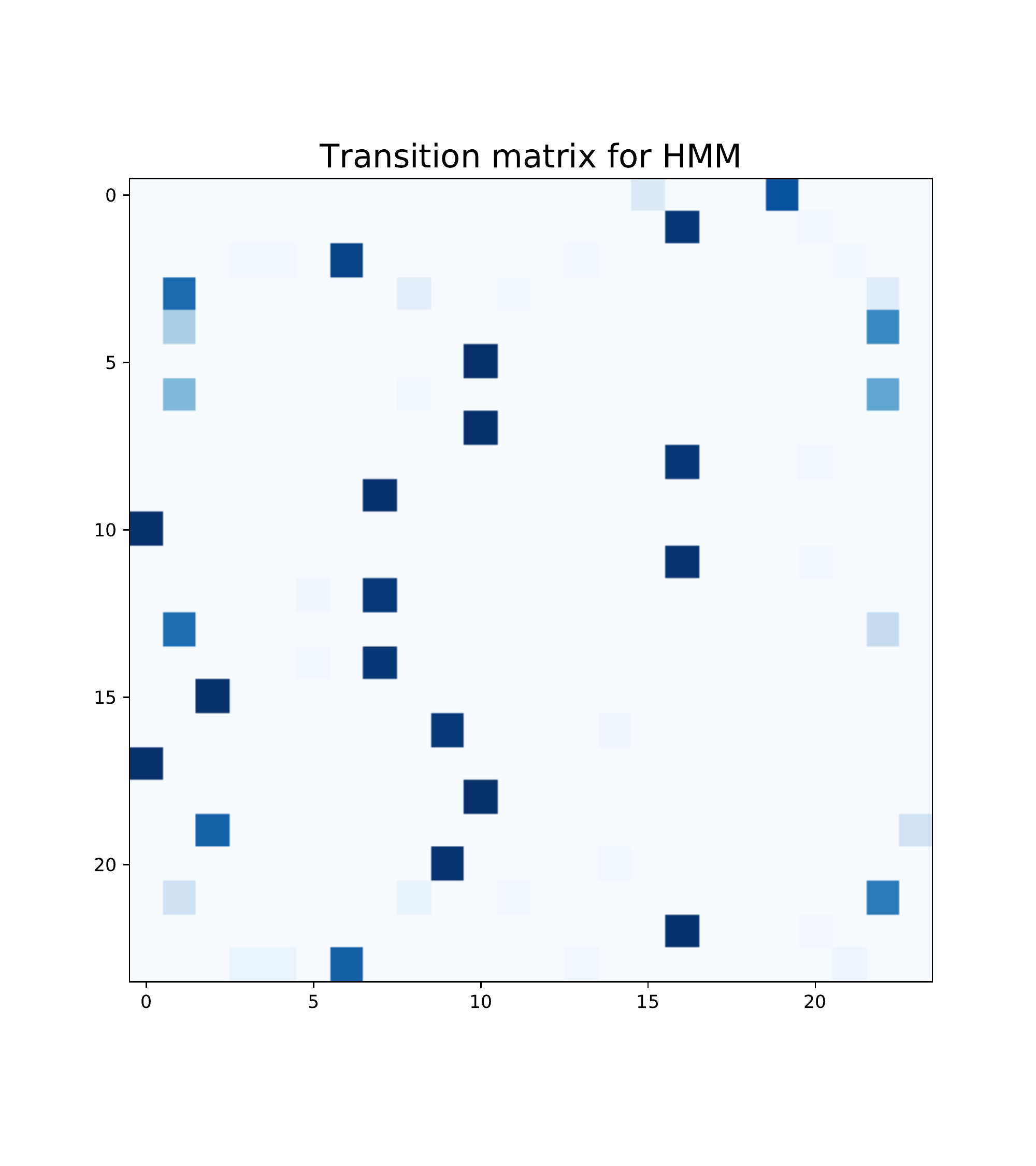}
    \vspace{-1cm}
    \caption{Overcomplete HMM}
    \label{fig:tmb}
\end{subfigure}
\caption{Learned transition matrix when trained on data generated from the toy example}
\label{fig:tm}
\end{figure}

Fig \ref{fig:tm}(a) shows the learned transition matrix when a CHMM is used to learn a model for the sequence generated by the toy FSM. The CHMM was initialized with just two clones for each symbol and it was able to exactly recover the underlying sequence generation structure and achieve the best possible predictive performance. Observe that the number of clones do not correspond to the temporal order: this example has eight times-steps of temporal dependency, but this is captured using only two clones. In contrast, a regular overcomplete HMM failed to capture the long-term dependence corresponding to the generating FSM, see Fig \ref{fig:tm}(b). The problem here is not that the overcomplete HMM is insufficient, but the opposite, its extra flexibility makes it converge to a bad local minimum.

\subsection{Related Work}
N-grams, when combined with Kneser-Ney smoothing \cite{baldi1994smooth}, are one of the strongest baselines for sequence modeling, being the tool of choice in many settings. Their weakness, however, is that they base their predictions exclusively on the statistics gathered over a finite context. When the required context is small this is not a problem, but since the model size grows exponentially with the length of the context, this quickly results in a) storage issues and b) insufficient data to fit a good model. For instance, in the above FSM, we can increase the length of the random sequence from 8 to an arbitrarily large number and the optimal model for the sequence will still be expressible as a CHMM with 2 clones. On the other hand, the context of the n-grams will grow arbitrarily large while memorizing all the exponentially many random sequences that are observed, becoming unfeasible at some point. Sequence memoizers \cite{sequence-memoizer} provide better smoothing mechanisms than n-grams, but share the same shortcoming as n-grams in not being able to efficiently represent sequences with ``holes''.

The idea of cloning the states of a Markov chain to create higher-order representations can be traced back to a popular compression algorithm \cite{cormack1987data}. Clones are created by identifying the states in a lower-order model that need to be split, and then relearning the counts. This basic idea has been re-discovered and elaborated. multiple times in different fields \cite{ostendorf1997hmm, hawkins2009sequence, xu2016representing,katahira2011complex, okubo2015growth}. Notably, cloned state representations are postulated to be the underlying mechanisms for representing sequential structure in bird songs \cite{katahira2011complex,okubo2015growth}. While splitting states is a viable method for learning simple sequential structure, it encounters problems with complex sequences (see Appendix \ref{sec:greedysplitting}). Partially observable Markov Models (POMMs) \cite{callut2005inducing} utilize the same idea of a cloned structure and recognize their connection to HMMs, but utilized a greedy clone splitting strategy to learn the HMM.

It is well recognized that constraining an HMM appropriately can alleviate the local minima problem of EM \cite{roweis2000constrained, hmm}. The only difference between the CHMM and the HMM is that in the CHMM the emission matrix is initialized to a particular deterministic sparse matrix which does not change with further learning. Even though this makes the CHMM less expressive compared to HMMs\footnote{For the same number of hidden states. An HMM can always be expressed as a CHMM by increasing the number of hidden states and embedding the emission matrix of the HMM in the transition matrix of the CHMM.}, it also makes it less prone to finding poor local minima. We show in Section \ref{sec:experiments} that the above FSM cannot be learned as accurately by an HMM as it is by a CHMM. Even more importantly, knowing the emission matrix a priori results in important computational savings, as we will show in Section \ref{sec:model}, making CHMMs more widely applicable. Overcomplete HMMs are appealing because of their ability to represent higher-order information, and \cite{overcomplete-hmm-nips} anticipates that specific sparsity structures on the transition matrix of an overcomplete HMM can lead to efficient learning. However, restricting the transition matrix to particular classes like left-to-right or cyclic can limit the expressive power of the model and limit its application. One, perhaps surprising, insight of CHMM is that imposing the sparsity structure on the emission matrix leads to a sparse transition matrix where the sparsity pattern is learned from data. 
 
RNNs, in their different incarnations, keep a state that is propagated while modeling the sequence, which means that they can properly model this type of sequence. The CHMM can be seen as the fully generative version of an RNN, thus being able to deal with all sorts of probabilistic queries in closed form.

\section{Learning a CHMM}
\label{sec:model}
Since the CHMM is just a normal HMM with additional restrictions, the same equations for learning HMMs apply, with some simplifications. These simplifications result in significant savings in compute and memory.

The probability of a sequence under this model can be expressed as the sum over the (exponentially many) hidden sequences that are compatible with the observed sequence. The probability of each hidden sequence is straightforward to compute using the chain rule:
\begin{equation}
P(x_1,\ldots,x_N) =  \sum_{\{z_n \in \text{hid}(x_n)\}_{n=1}^N} P(z_1) \prod_{n=1}^{N-1}  P(z_{n+1}|z_n)
\end{equation}
where $z_n \in \text{hid}(x_n)$ means that the summation is only over the hidden values of $z_n$ that emit $x_n$. There is a fixed, known association between hidden states and emissions, with each hidden state being associated to exclusively one emission. As noted in the previous section, the different hidden states associated to one emission are called the clones of that emission.

The parameters of this model are the prior probabilities, collected in the vector $\pi$, such that $\pi_u = P(z_1 = u)$, the transition probabilities $T$, such that $T_{uv} = P(z_{n+1} = v | z_n=u)$, and the number of clones allocated to each emission. Matrix $T$ is row stochastic, i.e, $\sum_v T_{uv} = 1$.

A simple and effective way to decide how to allocate clones to each symbol is to do so proportionally to the number of times that the symbol appears in the data. The intuition that more frequent symbols need more clones is obvious when we go to the extreme case of memorizing a sequence: we need exactly as many states as total symbols appear in the sequence, and as many clones per symbol as occurrences of that symbol appear in the sequence.

\paragraph{Expectation-Maximization for CHMM:} If the hidden states are ordered in such a way that all the clones of the first emission are placed first, all the clones of the second emission are placed second, etc, we can divide the transition matrix in submatrices, in which each submatrix is relevant for the transition between a pair of symbols. I.e., the transition matrix $T$ can be broken down into smaller submatrices $T(i, j)$ that contain the transition probabilities $P(z_{n+1} | z_n)$ such that $z_n\in i$ and $z_{n+1} \in j$. The subindices $i, j$ range in $1\dots E$, where $E$ is the total number of emitted symbols. I.e., they identify a specific set of clones. Arranging the $T(i, j)$ in a grid according to their subindices recovers the full transition matrix $T$. Similarly, the prior probability vector $\pi$ can be broken in sub-vectors $\pi(i)$ that contain the prior probabilities $P(z_1)$ such that $z_1 \in i$. Stacking all the sub-vectors, we recover $\pi$. The standard Baum-Welch equations can then be expressed in a simpler form:

\paragraph{E-Step}

\begin{align*}
\alpha(1) &= \pi(x_1) ~~&~~
\beta(N) &= 1(x_N)\\
\alpha(n+1)^\top &= \alpha(n)^\top T(x_n, x_{n+1}) ~~&~~
\beta(n) &= T(x_n, x_{n+1}) \beta(n+1)
\end{align*}

\paragraph{M-Step}

\begin{align*}
\gamma(n) &=  \frac{\alpha(n) \circ \beta(n)}{\alpha(n)^\top\beta(n)} ~~&~~
\xi_{ij}(n) &=  \frac{\alpha(n)\circ T(i, j) \circ \beta(n+1)^\top}{\alpha(n)^\top T(i, j) \beta(n+1)}\\
\pi(x_1) &= \gamma(1)  ~~&~~
T(i, j) &=  \sum_{n=1}^N \xi_{ij}(n) \oslash \sum_{j=1}^E \sum_{n=1}^N \xi_{ij}(n)
\end{align*}

where the symbol $\circ$ is the elementwise product (with broadcasting where needed) and $\oslash$ is the elementwise division (with broadcasting where needed). All vectors are column $M\times1$ vectors, where $M$ is the number of clones per emission\footnote{We use a constant number of clones per emission for simplicity here, but this number can be selected independently for each emission.}.

\paragraph{Online EM for CHMM:}  It is possible to develop an adaptive, online version of the above algorithm by splitting the sequence in $B$ contiguous batches $b = 1\ldots B$ and performing EM steps on each batch successively. We store our running statistic after processing batch $b$ in $A^{(b)}$, and compute $T^{(b)}$ from it. We have
$$
A_{ij}^{(b)} = (1-\lambda)\sum_{k=1}^b\lambda^{b-k}\sum_{n\in \text{batch}(k)} \xi_{ij}(n)~~~~~~ T^{(b)}(i, j) = A_{ij}^{(b)} \oslash \sum_{j=1}^E A_{ij}^{(b)},
$$
where $0<\lambda<1$ is a memory parameter and $n\in \text{batch}(k)$ refers to the time steps contained in batch $k$. If we set $\lambda \to 1$, then $T^{(B)}(i, j)$ would coincide exactly with the transition matrix from the previous section. For smaller values of $\lambda$, the expected counts are weighed using an exponential window\footnote{Note that proper normalization of the exponential window is unnecessary, since it will cancel when computing $T^{(b)}(i, j)$.}, thus giving more weight to the more recent counts. To obtain an online formulation, we simply express $A_{ij}^{(b)}$ recursively
$$
A_{ij}^{(b)} = \lambda A_{ij}^{(b-1)} + (1-\lambda)\sum_{n\in \text{batch}(b)} \xi_{ij}(n),
$$
so that the expected counts of the last observed batch are incorporated in our running statistic. The statistics $\xi_{ij}(n)$ of batch $b$ are computed from the E step over that batch, which uses the transition matrix $T^{(b-1)}$, obtained after processing the previous batch. This allows to learn from arbitrarily long sequences, and to adapt to changes in the statistics if those happen over time. If we are interested in online learning but not adaptivity, we can feed all the batches from the same sequence multiple times (analogous to the epochs in an NN), so that learning can continue to make progress after one full pass through the sequence.

\paragraph{Computational savings:} In a standard HMM with $H$ hidden states in which the emission matrix is stochastic, the computational cost for a sequence of length $N$ is ${\cal O}(H^2N)$ and the required storage is ${\cal O}(H^2 + HN)$, including the space for the transition matrix and the forward-backward messages.

In contrast, the CHMM exploits the known sparse emission matrix, which results in savings. With $H$ hidden states and $M$ clones per emission the computational cost is ${\cal O}(M^2N)$ and the memory requirement is worst case ${\cal O}(H^2 + MN)$, with additional savings for every pair of symbols that never appear consecutively in the training sequence (since the corresponding submatrix will be zero and does not need to be stored). Space requirements can be improved even further by using the online version of EM described above.

Since $H = ME$ where $E$ is the total number of symbols, an increase in alphabet size does not increase the computational cost of the CHMM, whereas it does increase the computational cost of the HMM. Since the transition matrix can be stored sparsely, storage cost may or may not be increased depending on the structure of the sequence.


\section{Empirical Results}
\label{sec:experiments}

We evaluate CHMMs on two classes of data sets that test a model's ability to learn long-term temporal structure: 1) Synthetic data sets, which serve to illustrate the model's representational properties, 2) Character-level language models, which demonstrate that CHMMs can learn on real data distributions. We show that CHMMs outperform LSTM-RNNs, n-grams and sequence memoizers in both the language modeling and synthetic data tasks. 

The metric by which these methods are compared is bits per symbol ---referred to as \textsc{BPS}--- which reflects the compressive capacity of a model. For a sequence $x=(x_1,\ldots,x_N)$ and a set of parameters $\Theta$, the \textsc{BPS} is defined as:
$$ \textsc{BPS} = - \frac{1}{N} \log_2 P(x| \Theta) = - \frac{1}{N} \big( \log_2 P(x_1 | \Theta) +  \sum_{n=2}^N \log_2 P(x_n | x_1, \ldots, x_{n-1}, \Theta) \big).$$

\subsection{Synthetic data sets}
We test CHMMs on two synthetic data sets: the toy example described in Section \ref{sec:example} and a sequence that opens and closes parenthesis and brackets in a syntactically correct way (see below). CHMMs outperformed standard overcomplete HMMs on both these tasks. On a more challenging version of these synthetic tasks, CHMMs outperformed LSTMs. Our experiments additionally reveal the computational advantage of using the online EM training procedure.

\subsubsection{Toy example}\label{sec:toy-example}

Data for the toy example is generated from a generalized version of the FSM presented in Section \ref{sec:example}. In that FSM, node $0$ deterministically led to node $1$ through a sequence of stochastically switching nodes in between. Node $2$ deterministically led to node $3$ in a similar manner. We keep the same structure but add the parameter $k$ to control the separation between $\{0,1\}$ and $\{2,3\}$. We define $0, 1, 2,$ and $3$ as \textit{signal} elements, which are separated by $3k - 2$ \textit{noise} symbols, with values in $\{4, \ldots 4k+1\}$.  We can interpret $k$ as the number of ``holes'' in the context. The example of  Section \ref{sec:example} corresponds to using $k=3$.

Train and test data are generated by concatenating stochastic sequences of length $3k$. Each stochastic \textit{noise} sequence of length $3k - 2$ is emitted by repeating the following loop -- indexed over $n$. For $n\ge 1$, the  symbol $4n$ and $4n + 1$ is emitted with $0.5$ probability. Symbol $4n+2$ is then deterministically emitted and followed by $4n+3$. Finally, elements $4(n+1)$ and $4(n+1) + 1$ are emitted with a $0.5$ probability each. The last two noise elements ($4k$ and $4k+1$) are  connected with one of the signal elements $2$ or $3$, depending on whether the last signal element emitted was $0$ or $1$. As discussed in Section \ref{sec:example}, this data can be drawn from a cloned HMM with $1$ clone for all signal elements and $2$ clones for the noise elements. We consider different training set sizes, as specified in Table \ref{table:synthetic}.

Note that even though in theory 2 clones are enough to model this sequence, in practice the learning algorithm might not find the optimal 2 clone model, i.e., it might converge to a bad local minimum. Increasing the number of available clones (thus using an over-flexible model for this sequence) is an effective way to get around the local minima problem.

\paragraph{Methods compared: } We compare the performance of a CHMM (with standard and online EM training) with an overcomplete HMM.
\begin{itemize}
    \item \textsc{CHMM trained with EM: } For $k \le 3$, the CHMM is initialized with the optimal number of clones for each element. We allocate an extra clone to each noise element for $k=4$, and an additional one for $k=5$.  The Baum-Welch algorithm presented in Section \ref{sec:model} is run for a maximum number of $\text{It}_{\max}=1000$ iterations. We stop the algorithm when the relative gain in train likelihood between 2 iterations of the algorithm is lower than $\epsilon = 10^{-6}$.
    \item \textsc{CHMM trained with online EM: } This is the same model as above, but we use the online EM algorithm. We split our training data into batches of size $400$ and consider a discount factor of $\lambda = 0.9$.
    \item \textsc{HMM: } This is an overcomplete HMM with same number of hidden states than the cloned HMMs above: $8k$ for $k\le3$, $46$ for $k=4$ and $76$ for $k=5$. We use Baum-Welch algorithm from the \textsc{hmmlearn} package\footnote{Available in \textsc{scikit learn}  \cite{scikit-learn}.}. We use the same transition matrix initialization (random stochastic), number of iterations and convergence criterion as for the CHMMs.
\end{itemize}


The results of the comparison tabulated in Table \ref{table:synthetic} show that CHMM outperforms HMM on this data set. Training with online EM results in better models and faster convergence compared to standard EM. CHMMs trained with online EM is able to achieve  almost optimal performance on $k=2,3,$ and $4$. For $k=5$, it reaches the optimal performance for a  portion of the random initializations.


\begin{table*}[!h]
	\centering
	\caption{Comparison of CHMM with an overcomplete HMM on the toy example. HMM converges to a local minimum and never learns the optimal transition and emission matrices. CHMM recovers the optimal transition matrix with regular EM training for $k=2$. The online EM algorithm accelerates convergence with more frequent updates of the transition matrix. As a result, it systematically converges to an optimal solution for $k\le4$, and in some runs for $k=5$. BPS values are averages over 10 runs and standard deviations are noted in parentheses.}
	\bigskip
	\label{table:synthetic}
	\resizebox{\textwidth}{!}{
		\begin{tabular}{lllllllllll}
			\cmidrule{4-10}
			&&&\multicolumn{2}{c}{\textsc{CHMM with EM}} & \multicolumn{2}{c}{\textsc{CHMM with online EM}} & \textsc{HMM} & \multicolumn{2}{c}{\textsc{Optimal}} \\
			\toprule
			$k$ & Training size & Max clones  & BPS & N iter & BPS & N iter & BPS & BPS & Clones  \\
			\toprule
			2 & 11250 & 2 & 0.502 (0.002)  & 71.4 (13.6) & 0.502 (0.001) & 11.9  (0.7) & 0.657 (0.000) & 0.500 & 2 \\
			3 & 33750 & 2 & 0.556 (0.000) & 42.9 (73.2) & 0.446 (0.002) & 94.1 (37.1) & 0.556 (0.000) & 0.444 & 2 \\
			4  & 90000 & 3 & 0.500 (0.000) & 50.4 (27.5) & 0.418 (0.000) & 284.4 (212.4) & 0.500 (0.000) & 0.417 & 2\\
			5 & 225000 & 4 & 0.467 (0.000) & 1000.0 (0.0) & 0.428 (0.033) & 732.0 (240.0) & 0.467 (0.000)  & 0.400 & 2\\
			\bottomrule
		\end{tabular}
	}
\end{table*}

\subsubsection{Bracket example}

A sequence in this data set consists of brackets and parenthesis [, (, ), ] forming syntactically valid sequence. The syntax is the expected: they must appear in opening-closing pairs, with both elements matching in type and nesting. Sentences are terminated with the symbol $\vert$.
As an example, a syntactically correct sequence with maximum nesting $k=3$ is: $\vert ( [ ] [ ( ) ] ) ( ( ) ( ) (  ) ( [ ] ) [ ] ( ) ) ( ( [ ] ) [ ] ) \vert [ [ ] ] \vert ( ( ) ) \vert [ ] \vert$. 

As in the above example, we compared two variants of CHMM training with an overcomplete HMM. We could only train the HMM for a few tens of iterations -- which makes it comparable to both variants of CHMM -- because of convergence issues. CHMM achieves the optimal \textsc{BPS} for $k=2$ with batch EM and online EM, and for $k=3$ only with online EM. We additionally found that for $k=2$, online EM can solve the problem with $4$ clones whereas batch EM needs at least $6$ clones.  An overcomplete HMM is unable to achieve these results in either setting. 

\begin{table*}[!h]
	\centering
	\caption{Comparison of HMM and CHMM on the bracket example.}
	\bigskip
	\label{table:bracket}
	\resizebox{\textwidth}{!}{
		\begin{tabular}{lllllllllll}
			\cmidrule{4-10}
			&&&\multicolumn{2}{c}{\textsc{CHMM with EM}} & \multicolumn{2}{c}{\textsc{CHMM with online EM }} & \textsc{HMM} & \multicolumn{2}{c}{\textsc{Optimal}} \\
			\toprule
			$k$ & Training size & Max clones  & BPS & N iter & BPS & N iter & BPS & BPS & Clones  \\
			\toprule
			2 & 50000 & 6 & 1.053 (0.003)  & 37.2 (1.0) & 1.054 (0.02) & 5.6 (0.5) & 1.178 (0.004) & 1.053 (0.003) & 3 \\
			3 & 50000 & 20 & 1.172 (0.003) & 69.2 (1.3) & 1.134 (0.002) & 26.7 (2.1) & 1.384 (0.005) & 1.130 (0.003) & 7 \\
			3 & 50000 & 100 & 1.191 (0.004) & 47.6 (12.4) & 1.240 (0.002) & 99.9 (0.3) & 1.500 (0.006) & 1.161 (0.004) & 15 \\
			\bottomrule
		\end{tabular}
	}
\end{table*}

\subsubsection{Comparison with LSTM}

We introduce two minor modifications of the previous examples that increases the difficulty of the data sets. Our train and test distributions are now a function of some parameter $\alpha$. We show that CHMM is still able to extract relevant information from this more complex environment that allows it to generalize well. In comparison, we show that an LSTM is unable to recover such accurate information and leads to worse predictive performance. 

\paragraph{Toy example: } We fix $k=2$ and introduce a parameter $\alpha \in (0,1)$ which changes the generative model: the data is now generated from the FSM in Fig \ref{fig:plots}(a). The length of the noise is now stochastic, and its expected length varies with $\alpha$: a large $\alpha$ will favor longer sequences -- note that the results described earlier consider $\alpha=0$. We now fix $\alpha_u = 0.9$, $\alpha_d = 0.1$ and generate the train and test sets as mixture of sequences with $\alpha = \alpha_u$ and sequences with $\alpha = \alpha_d$, where $P(\alpha = \alpha_u) = 0.95$. Our train and test sequences have the same distribution and are composed of a large majority of long sequences.

\paragraph{Bracket example: }
We consider the bracket example with maximum nesting $k=2$. We define $\alpha^0_{(}$ and $\alpha^0_{[}$ as the respective probabilities of starting a sentence with a parenthesis and a bracket. Similarly, we introduce the probabilities $\alpha^1_{(}$, $\alpha^1_{[}$ and $\alpha^1_{),]}$ for the first degree of nesting -- a parenthesis or a bracket has already been opened -- of respectively opening a parenthesis, opening a bracket or closing the previously opened symbol. 

Our previous setting considered $\alpha^0_{(}=\alpha^0_{[}$ and $\alpha^1_{(} = \alpha^1_{[} = \alpha^1_{),]}$. We generate the training sequence with $\alpha^0_{(}= \frac{15}{19}, \alpha^0_{[} =\frac{4}{19} $ and $\alpha^1_{(} = \frac{15}{24},  \alpha^1_{[} = \frac{4}{24}$, and $\alpha^1_{),]} = \frac{5}{24}$. When generating the test sequence, we set $\alpha^0_{(}= \frac{4}{19}, \alpha^0_{[} =\frac{15}{19} $, $\alpha^1_{(} = \frac{4}{24},  \alpha^1_{[} = \frac{15}{24}$, and $\alpha^1_{),]} = \frac{5}{24}$.

\paragraph{Results: } The following Table \ref{table:lstm-chmm} compares the methods:
\begin{itemize}
    \item \textsc{CHMM: } Cloned HMM described trained with online EM.
	\item \textsc{LSTM: } We use the LSTM from the popular repository \url{https://github.com/karpathy/char-rnn} trained with default parameters: the number of layers is 2 and the hidden size 100. We set the maximum number of epochs to 400. We generate an additional validation set with same distribution as the test set, which is used to assess convergence. 
\end{itemize}
\begin{table*}[!h]
	\centering
	\caption{Comparison of cloned HMM and LSTM on the modified toy and bracket example described above. The model-based cloned HMM captures the longer term dependencies of the data whereas LSTM fails.}
	\bigskip
	\label{table:lstm-chmm}
	\begin{tabular}{llllllll}
		\cmidrule{3-7}
		&&\multicolumn{3}{c}{\textsc{CHMM with online EM}} & \multicolumn{2}{c}{\textsc{LSTM}} \\
		\toprule
		Data set & Avg training set size & Max clones & BPS & N iter & BPS & N iter \\
		\toprule
		Toy & 59559 & 2 & $\B{0.512}$ (0.003)  & 91.9 (53.3) & 0.535 (0.010) & 400\\
		Bracket & 50000 & 4 & $\B{1.191}$ (0.015)  & 15.9 (13.8) & 1.211 (0.012) & 400\\
		\bottomrule
	\end{tabular}
\end{table*}

\subsection{Real data sets}

We evaluated CHMMs on character-level language modeling with eight text data sets available online:
\begin{itemize}
	\item Six online books accessible from \textsc{Python}'s Natural Language Toolkit package data sets \cite{nltk}: \textsc{blake-poems, caroll-alice, shakespeare-hamlet, shakespeare-macbeth, milton-paradise} and \textsc{melville-mobydick}.
	\item The \textsc{war-peace} data set accessible from the \textsc{LSTM} repository above.
	\item \textsc{book1} from the \textsc{calgary} repo, available at: \url{https://www.ics.uci.edu/~dan/class/267/datasets/calgary/?C=M;O=D}
\end{itemize}

For each data set, we first transform all numbers to their character forms and replace all words which appear only once in the training or only in the test set into the word \textit{rare}. We then remove all characters other than the 26 letters and space. The first 90\% of our data is used as training set, and the remaining 10\% as the test set. In addition, for computational reason, we limit our training set size to $750,000$: this enforces us to reduce the training set ratio on some data sets. 

\paragraph{Methods compared: } We compare CHMM with other commonly used methods for character-level language models:
\begin{itemize}
	\item \textsc{CHMM: } We train CHMMs using the sparse EM algorithm described in Section \ref{sec:model}. Smoothing is known to be critical for the sequence prediction task \cite{smoothing-benchmark} to avoid overfitting. We regularize via early-stopping \cite{yao2007early, fleming1990equivalence}. A total capacity of $30000$ states are allocated to characters, proportional to the number of unique n-grams that end with that character. $10\%$ of the training sequence  is retained for validation. The training procedure consists of $\text{It}_1$ iterations of online EM and $\text{It}_2$ iterations of standard EM, where these parameters are selected using the validation set. After selecting the early stopping parameters, we retrain on the entire training and validation set.
	\item \textsc{SeqM: } Hierarchical Bayesian model described in \cite{sequence-memoizer}. The authors released an online implementation accessible at \url{https://github.com/jgasthaus/libPLUMP}. We tune the smoothing parameter over the same validation dataset used for the CHMM and then retrain on the entire data set.
	\item \textsc{n-grams: } We use n-grams with Kneser-Ney smoothing \cite{kneser-ney} as implemented in the Kyoto Language Modeling Toolkit \cite{kylm}. The code is accessible at \url{https://github.com/neubig/kylm} 
	\item \textsc{LSTM:} We compared against two LSTM implementations and report the best of the two results for each data set. The first is Karpathy's char-rnn mentioned above, and the second is a PyTorch version at \url{https://github.com/spro/char-rnn.pytorch}.
	
\end{itemize}

\paragraph{Results: } CHMMs outperformed the other methods, including LSTM, for most of the data sets we compared against, see Table \ref{table:real}. Although sequence memoizers consistently returned a better model compared to n-grams, they were significantly outperformed by the CHMM on all data sets except \textsc{WAR-PEACE}, on which it learned a slightly better model. 

Since LSTMs are the gold standard in character-level language models \cite{Karpathy}, we were surprised that CHMMs outperformed them. We searched over combinations of parameters of LSTM, including the number of layers and the size of the layers and were unable to find a setting that beat CHMMs on these data sets. One possibility is that LSTMs require significantly larger training data sets compared to the ones we use here. 

\begin{figure}[t!]
\begin{subfigure}{0.5\textwidth}
	\centering
    \includegraphics[width=3in]{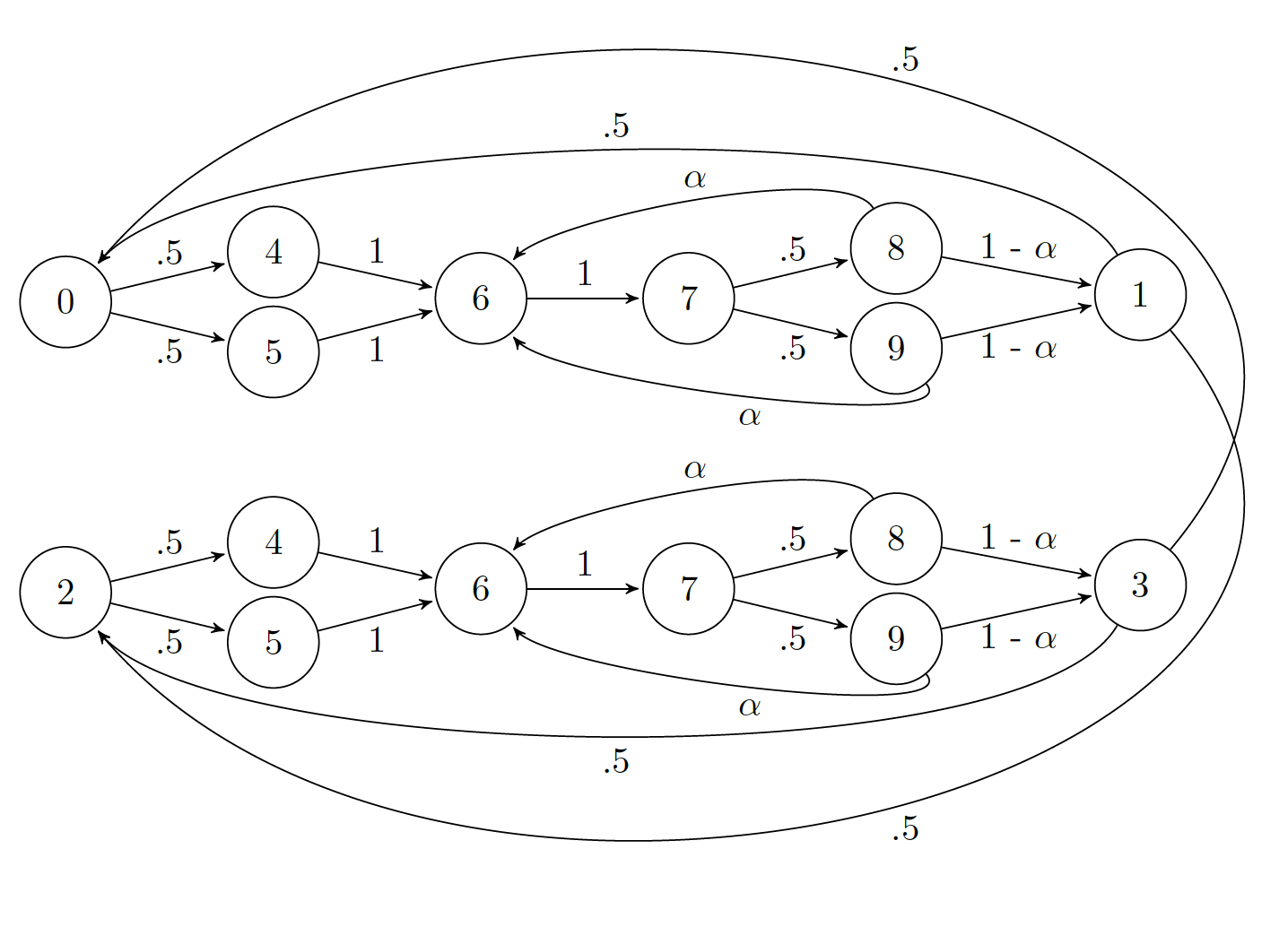}
	\caption{ Toy FSM with more complex transitions.}
\end{subfigure}
\begin{subfigure}{0.5\textwidth}
    \centering
    \includegraphics[width=3in]{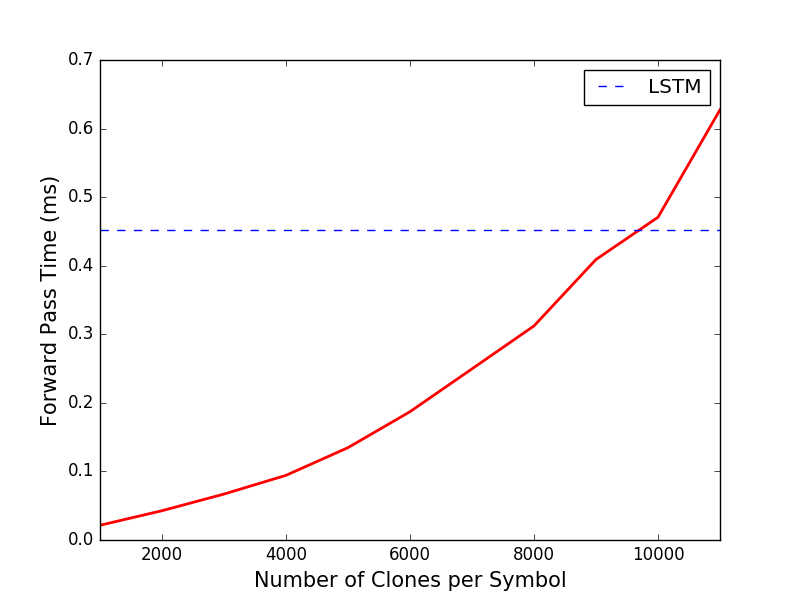}
    \caption{Forward pass computation time of CHMM vs LSTM.}
    \label{fig:chmm_timing}
\end{subfigure}
\begin{subfigure}{0.5\textwidth}
    \centering
    \vspace{-.4cm}
    \includegraphics[width=3in]{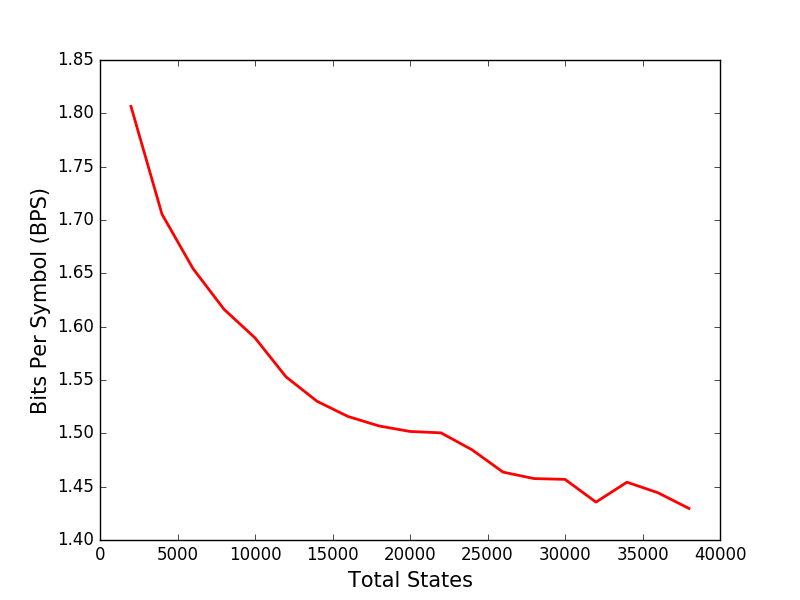}
    \caption{Training set performance of CHMMs.}
\end{subfigure}
\begin{subfigure}{0.5\textwidth}
    \centering
    \includegraphics[width=3in]{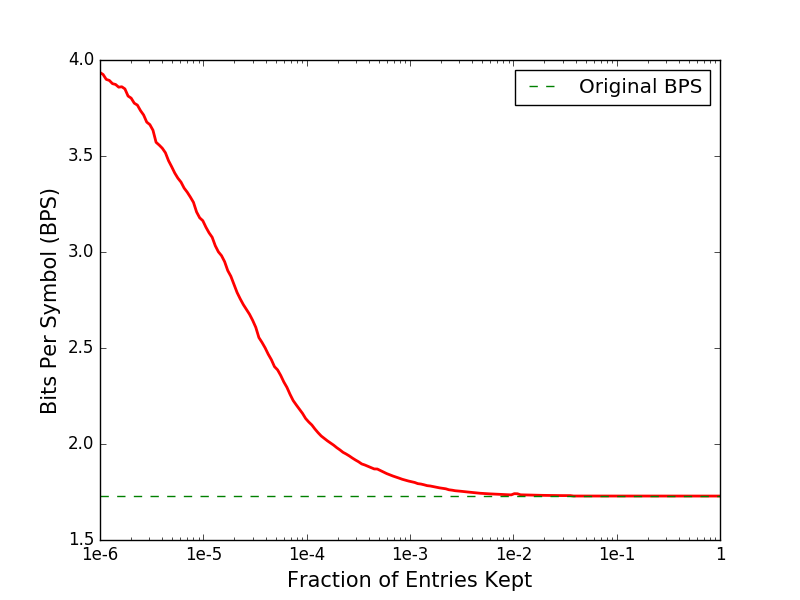}
    \caption{Effect on CHMMs performance of zeroing small values of the transition matrix.}
\end{subfigure}
\caption{(a)  Toy FSM with a more intricate temporal dependency used for comparison with LSTMs. (b) Comparison of forward pass computation time of CHMMs with LSTM. (c) Training set performance of CHMMs improve with increased capacity. (d) Close to 99 \% of the transition matrix entries can be set to zero without affecting CHMM's performance.}
\label{fig:plots}
\end{figure}

In our tests on character-level models, we found CHMMs to be significantly more computationally efficient than LSTMs during inference. For the forward pass inference on a sequence length of $42,297$, the optimal CHMM takes only $1.6$ seconds. In comparison, a PyTorch implementation of a one-layer LSTM with 256 hidden units takes $32.4$ seconds on the same GPU. Fig \ref{fig:plots}(b) shows the average time for a single step of inference in a CHMM as a function of the number of clones. 

The models that we train are HMMs with a very large state space and a large number of parameters. The character-level language models have $0.9$ billion parameters. After training, a large fraction of the parameters go to zero, or can be set to zero without affecting the performance. The resultant transition matrix is extremely sparse with ~99\% of the parameters being zero, as shown in Fig \ref{fig:plots}(d) for the Milton data set. Our GPU implementation of CHMM does not exploit this sparsity of the learned transition matrix, and additional computational speed advantage over the LSTM  could be gained by a sparse implementation.

Increasing the number of hidden states in a standard HMM does not always result in a better likelihood on the training data due to credit diffusion \cite{bengio1995diffusion} and local minima. In contrast, in our experiments we found that CHMMs with more clones consistently produced better training likelihoods compared to the ones with fewer clones, supporting the hypothesis that the sparsity structure imposed on the emission matrix serves to alleviate the credit diffusion problem. This obviously doesn't necessarily translate into an immediate improvement in test data performance when the training dataset is not large enough for the capacity of the model. Figure \ref{fig:plots}(c) shows the performance of a CHMM on the training data set as the capacity of the model is increased from $1,000$ to $50,000$ total states. All the results in Table \ref{table:real} were obtained using the same capacity of $30000$ hidden states for the CHMM. However, it is possible that their performance could be further improved by using the validation set to select an optimal capacity.

\begin{table*}[!h]
	\centering
	\caption{Bits per symbols comparison of cloned HMM versus n-grams and Sequence Memoizer on real text data sets. Cloned HMM reaches a significantly lower BPS with respect to its high-order Markov (n-grams) and hierarchical Bayesian model opponents.}
	\bigskip
	\label{table:real}
	\begin{tabular}{lllllll}
		\toprule
		Data set & Train set size & Test set size & \textsc{cloned HMM}  & \textsc{n-grams} & \textsc{SeqM} & \textsc{LSTM}  \\
		\toprule
		\textsc{blake-poems} & 29912 & 3300 & $\B{1.60} $ & 1.75 & 1.71 & 1.68 \\
		\textsc{calgary} & 638677 & 7116 & $\B{1.63}$ & 1.72 &  1.64 & 1.67 \\
		\textsc{caroll-alice} & 118931 & 13063 & $\B{1.54}$ & 1.61 &  1.57 & 1.58 \\
		\textsc{shakespeare-hamlet} & 130101 & 14332 & $\B{1.63}$ & 1.72 &  1.69 & 1.68\\
		\textsc{shakespeare-macbeth} & 79646 & 8824 & $\B{1.69}$ & 1.79 &  1.77 & 1.74\\
		\textsc{milton-paradise} & 382942 & 42297 & $\B{1.73}$ & 1.83 &  1.78 & 1.78\\
		\textsc{melville-mobydick} & 750000 & 387864 & $\B{1.72}$ & 1.81 &  1.73& 1.76 \\
		\textsc{war-peace} & 750000 & 2237883 & 1.59 & 1.65 &  $\B{1.57}$ & 1.62\\
		\bottomrule
	\end{tabular}
\end{table*}

\subsubsection{Discovering `word graphs' using community detection}

\begin{figure}[b!]
    \centering
    \includegraphics[width=5.25in]{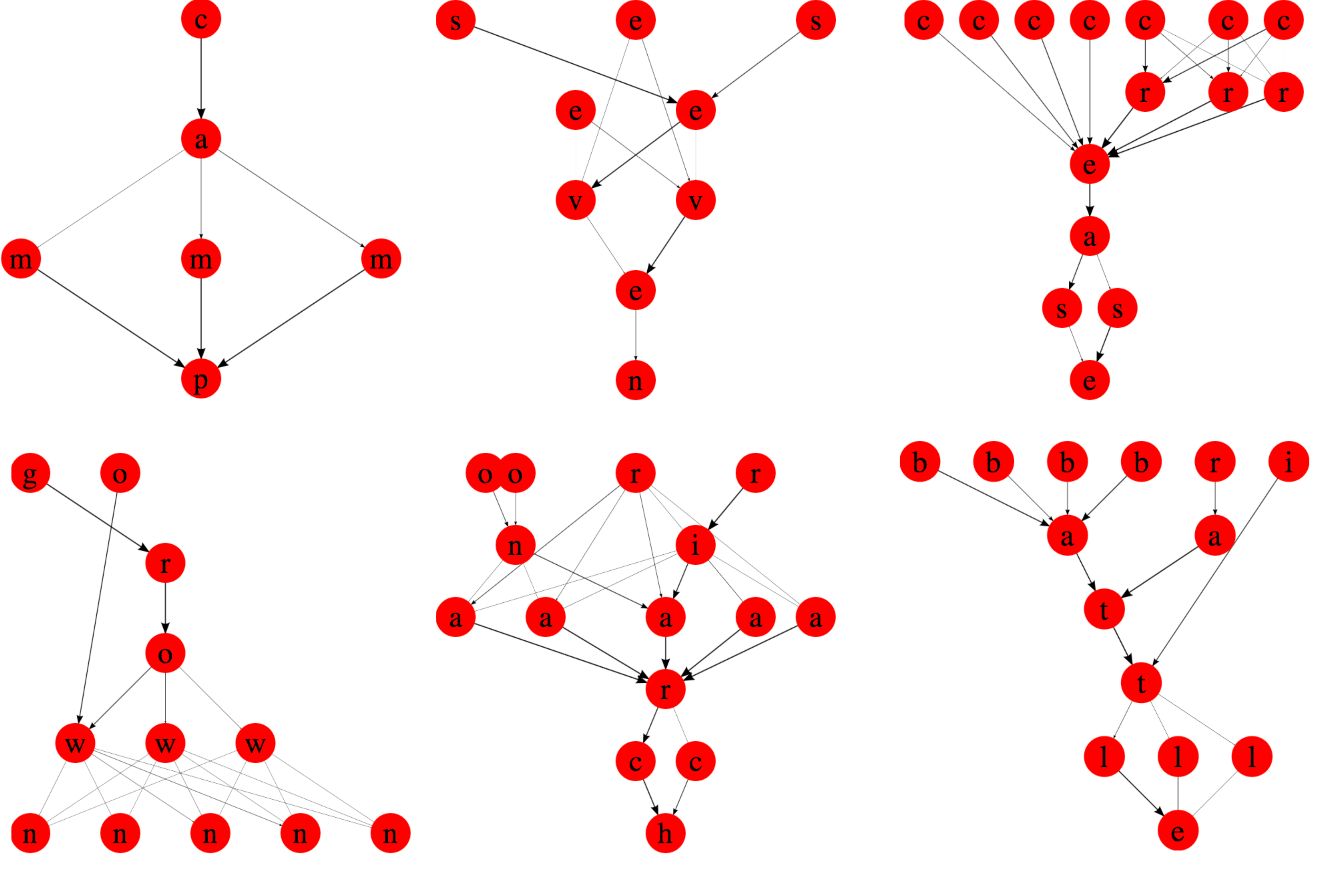}
    \caption{Examples of community subgraphs discovered in the CHMM transition matrix. Some
    communities correspond exclusively to frequent words or word segments. Many communities
    share multiple words with shared parts.}
    \label{fig:subgraphs}
\end{figure}

The transition matrix of a CHMM is a directed graph. Community detection algorithms \cite{rosvall2008maps} can be used to partition this graph into subgraphs to reveal the underlying structure of the data. 
For a CHMM trained on characters, the subgraphs discovered by Infomap-based community detection \cite{rosvall2008maps} correspond to words or word parts. Frequent and short words or word segments in the data get their own exclusive subgraphs, whereas infrequent and longer words share their graph representations with other words that have shared substrings. A community subgraph of a character-level CHMM can be thought of as a ``generalized word''. Some of those subgraphs have deterministic word productions based on what was seen in the training data, whereas some other sub-graphs produce different word-like combinations, many of which are completely novel.  

Figure \ref{fig:subgraphs} shows a few of the communities discovered in the CHMM transition matrix for \textsc{milton-paradise}, using the Infomap algorithm \cite{rosvall2008maps}. The bottom-right graph produces the words ``battle'' and ``rattle''. Interestingly, the word ``rattle'' does not appear in the training set, and is a generalization produced by CHMM. This generalization occurs by sharing the clone of ``t'' between ``battle'' and words like ``grateful'', ``generate'', etc. that have the ``rat'' segment in them.  

\begin{figure}[b!]
    \centering
    \includegraphics[width=5.25in]{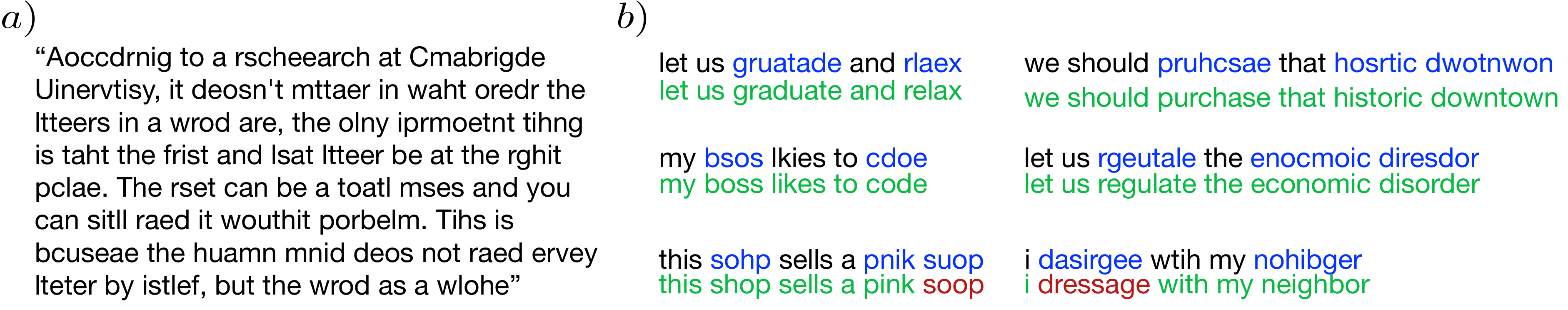}
    \caption{a) An internet meme about the ability of humans to decode scrambled text. Although the effect is true, the mentioned study is a hoax. b) Examples of sentences created from unseen words (in blue), with their CHMM Viterbi decodings shown in green. Errors are indicated in red.}
    \label{fig:scrambled_text}
\end{figure}

\subsubsection{Handling uncertainty}
As a demonstration of CHMM’s ability to handle uncertainty, we consider the problem of decoding scrambled text. People have a remarkable ability to decode scrambled text, a topic that has been an internet meme (Fig \ref{fig:scrambled_text}a): ``Aoccdrnig to a rscheearch at Cmabrigde Uinervtisy, it deosn't mttaer in waht oredr the ltteers in a wrod are, the olny iprmoetnt tihng is taht the frist and lsat ltteer be at the rghit pclae. The rset can be a toatl mses and you can sitll raed it wouthit porbelm. Tihs is bcuseae the huamn mnid deos not raed ervey lteter by istlef, but the wrod as a wlohe'' \cite{starling2018cna}. CHMMs can handle this without having to train on scrambled text if the input is encoded to reflect the uncertainty in the data. We treat the evidence for each in-between character (characters that are between the start of the word and end of the word) as uncertain, with the uncertainty uniformly distributed among the in-between characters of that word. With this encoding, performing MAP inference (Viterbi decoding) decodes the correct words in most cases.

We performed two experiments to test the ability of CHMMs to decode scrambled text. In the first test, we sampled thirty 600-long character sequences from \textsc{melville-mobydick}, and scrambled the words while preserving their beginning and ends. Using the uncertainty encoding described above, and using the CHMM that was trained on \textsc{melville-mobydick}, MAP inference decoded the sequences with 99.02\% word accuracy and 99.42\% character accuracy. The ability of CHMM to decode scrambled text is not limited to known words. To test this, we selected 638 unseen words from a list of 3000 most frequent English words and scrambled them as above. Out of these the 89 four-letter words were decoded with 74.1\% accuracy, the 99 five-letter words with 45.45\%, the 144 six-letter words with 40.3\%, the 172 seven-letter words with 27.9\%, and the 134 eight-letter words with 21.6\% accuracies. Figure \ref{fig:scrambled_text} shows examples of sentences created from the unseen words and their CHMM decodings. 

It is noteworthy that this capability is not using a separate dictionary, and not expecting the  words that are being decoded to be present in the training set of the CHMM. For example, the words “surprisingly” and “complementary” do not occur in the training set of the CHMM for the Moby Dick dataset. Nevertheless, the learned representation assigns non-zero likelihoods to those words, and their scrambled versions are corrected during inference. Doing this using an LSTM would require sampling from the LSTM and using a dictionary to identify correct words, whereas CHMM's internal representation and its ability to handle uncertainty allows it to solve this problem as a standard inference query. While this ability is similar to that of humans, we are not making the claim that humans use the same mechanism as CHMMs when they decode scrambled text.

\section{Theoretical Results}
This section summarizes our theoretical findings for CHMM: we refer to Appendix \ref{sec:guarantees} for a complete analysis. In particular, our next Theorem \ref{main-shrunk} presents statistical guarantees of the Baum-Welch algorithm for CHMM. It derives a finite sample L2 estimation of the difference between the successive Baum-Welch estimates $\{ \hat{T}_{\textsc{CHMM}}^j \}_{j\ge 0}$ and a transition matrix $T^*$, defined as a global optimum of the population likelihood -- the limit of the empirical likelihood in the case of an infinite amount of data. 

\begin{theorem}\label{main-shrunk}
	Let $x=(x_1,\ldots, x_N)$ be a sequence drawn from a CHMM. We assume Assumptions \ref{asu1} and \ref{asu2} hold -- as presented in Appendix \ref{sec:guarantees}. Then, we can fix $\gamma_{\textsc{CHMM}} < 1$, an integer $k>0$, and a sequence $r_{\textsc{CHMM}}(N,k, \delta)\tolim \limits_{N, k \to \infty} 0$ such that the sequence of CHMM Baum-Welch  iterates $\{ \hat{T}_{\textsc{CHMM}}^j \}_{j\ge 0}$ satisfy with probability at least $1 - \delta$:
	\begin{equation}\label{result-CHMM}
	\| \hat{T}_{\textsc{CHMM}}^j - T^*\|_2 \le \gamma_{\textsc{CHMM}}^j \| \hat{T}_{\textsc{CHMM}}^0  - T^* \|_2 + \frac{1}{1- \gamma_{\textsc{CHMM}}} r_{\textsc{CHMM}}(N,k, \delta), \;\;\ j\ge 0.
	\end{equation}
\end{theorem}

The L2 norm of the difference between the optimum $T^*$ and the Baum Welch iterates is bounded by the sum of an exponentially decaying term and a residual error. The residual is a function of the number of samples $N$ and the degree of truncation $k$ introduced in Appendix \ref{sec:guarantees}. Consequently, Theorem \ref{main-shrunk}   guarantees the local convergence of the Baum-Welch algorithm to a small ball around a global optimum of the population likelihood. The following Theorem \ref{comparison} proposes a similar bound for HMM and compares the L2 estimation upper bounds for CHMM and HMM.

\begin{theorem}\label{comparison}
	Under stronger assumptions than Theorem \ref{main-shrunk}, the sequence of transition matrices $\{ \hat{T}_{\textsc{HMM}}^j \}_{j\ge 0}$ derived via Baum-Welch algorithm for HMM satisfies a bound similar to Equation \eqref{result-CHMM} with probability at least $1-\delta$:
	\begin{equation}\label{result-HMM}
	\| \hat{T}_{\textsc{HMM}}^j - T^*\|_2 \le \gamma_{\textsc{HMM}}^j \| \hat{T}_{\textsc{HMM}}^0  - T^* \|_2 + \frac{1}{1- \gamma_{\textsc{HMM}}} r_{\textsc{HMM}}(N,k, \delta), \;\;\ j\ge 0.
	\end{equation}
	It additionally holds: $\gamma_{\textsc{CHMM}} < \gamma_{\textsc{HMM}}$ and $r_{\textsc{CHMM}}(N,k, \delta) < r_{\textsc{HMM}}(N,k, \delta)$.
\end{theorem}
Theorem \ref{comparison} guarantees the local convergence of the Baum-Welch algorithm for HMM. For fixed values of $N$ and $k$, the L2 estimation upper bound is larger for HMM than for CHMM. Two  major differences appear when both bounds are satisfied: (a) CHMM needs fewer iterations to converge to the local optima since its convergence rate -- of the order of $O( \gamma_{\textsc{CHMM}}^j )$ -- is smaller than the rate for HMM and (b) CHMM converges closer to the global optima since its radius of local convergence $r_{\textsc{CHMM}} / (1 - \gamma_{\textsc{CHMM}}) $ is smaller.

\section{Discussion}
\label{sec:discussion}
Learning the parameters of even simple graphical models like HMMs can be very challenging. In this work, we used ideas from neuroscience to constrain the structure of HMMs and showed that this results in efficient learning of temporal structure. The learned models can deal with uncertainty, a critical requirement for agents operating in a noisy and uncertain world. 

Although CHMMs outperformed LSTMs in many of our current settings, we would expect LSTM's performance to be superior when trained on longer datasets. Also, we know that LSTMs can handle large vocabularies well, while this remains to be shown for the case of CHMMs. Despite these caveats, CHMMs seem a good model for sequential structure learning when the generative structure needs to be understood, when this structure is relatively sparse, and when uncertainty needs to be handled. It is also noteworthy that some of the most modern deep learning language models can only handle finite contexts and would not be able to deal with the arbitrarily long memories required by some of the sequences that a CHMM can model.

Our work has opened up several directions for future research. Although the discovered matrices are very sparse, EM has a tendency to overuse the clones. Our preliminary investigations show that imposing a sparse prior (e.g. a Dirichlet prior) on the transitions in a variational Bayesian formulation results in further sparsity. 
CHMMs induce generalization using a capacity bottleneck on the clones. Further generalization could be obtained using appropriate smoothing mechanisms. One fruitful avenue could be to investigate how n-gram smoothing ideas can be transferred to the CHMM setting. Since the representational ideas of CHMMs originated in biology, investigating the mapping between sequential representations in the neocortex and CHMMs would be an interesting avenue of future research.

\section*{Acknowledgments}
We thank Michael Stark, Bruno Olshausen and Tom Silver for helpful comments on the manuscript.

\appendix
\section*{Appendix}

\section{Greedy clone splitting}
\label{sec:greedysplitting}
Here we show how greedy clone splitting can fail even in very simple cases. Consider a sequence formed by successively concatenating either ``ab'' or ``(ab)'', choosing between the two at random. An example sequence would be ``abab(ab)ab(ab)(ab)abab''. Sequences generated in this fashion can be modeled exactly as a CHMM with one clone for the symbols \emph{(} and \emph{)}, and two clones for the symbols \emph{a} and \emph{b} (one clone activates inside the parenthesis, the other outside). A model with a single clone for every symbol will obviously have a lower likelihood given the data.

Using Baum-Welch, the optimal model is readily found from training data when enough clones are available. Now consider a greedy clone-splitting mechanism that starts with one clone per symbol and splits one clone at a time when doing so increases the likelihood of the model:

\begin{itemize}
    \item Splitting \emph{a}'s clone cannot increase the likelihood of the model, since the single existing clone already predicts the single clone of \emph{b} with probability 1 and is always correct in doing so. 
    \item Splitting \emph{b}'s clone has the symmetric problem: even though having a different clone activate inside and outside the parenthesis would increase its ability to predict the next symbol (for instance, it should always be \emph{)} for the clone inside the parenthesis), since it is always preceded by the same clone of \emph{a}, it is not possible to disambiguate both clones of \emph{b}, and therefore this split does not increase the likelihood of the model either.
\end{itemize}
 
 In this toy example, the only move that results in a likelihood increase is one that splits the clones of \emph{a} and \emph{b} simultaneously, and not greedily.

\section{Theoretical guarantees}
\label{sec:guarantees}

\subsection{Upper bounds for CHMM}\label{sec:ub-chmm}
We consider $\alpha > 0$ , a sequence ${x}=(x_1,\ldots, x_N)$ with corresponding latent variables (clones) ${z}=(z_1,\ldots, z_N)$  drawn from a clone HMM with transition matrix ${T}_0$. The clones takes their values in $1,\ldots K$ and satisfy the following assumption:
\begin{asu}\label{asu1}
The Markov chain associated to the latent sequence is aperiodic, recurrent, has unique stationary distribution $\pi(.|{T}_0)$, and is revertible. In addition, we can fix a constant $\epsilon \in [\alpha, 1]$ such that:
\begin{equation}
\epsilon \pi(z_n | {T}_0) \le \mathbb{P}(z_n | z_{n-1}, {T}_0) \le \epsilon^{-1} \pi(z_n | {T}_0), \;\; \forall z_n, z_{n-1}.
\end{equation}
\end{asu}
Assumption \ref{asu1} is standard in the literature \cite{yang2017statistical}. Using Jensen's inequality, the normalized log-likelihood of the complete data $(z,x)$ is lower-bounded by:
$$ \frac{1}{N} \log\left( \sum_{{z}} \mathbb{P}({z}, {x} | {T}_0) \right)\ge Q_N({T}_0 | \tilde{{T}} ) := \frac{1}{N} \mathbb{E}_{ {z} | {x}, \tilde{{T}} } \left[ \log \mathbb{P}( {z}, {x}| {T}_0 ) \right], \;\; \forall \tilde{{T}} \in \Omega,$$
where $\Omega$ is the set of admissible transition matrices. The empirical operator $Q_N$ for clone HMM can be expressed as:
$$Q_N({T}_0 | \tilde{{T}} ) = \frac{1}{N} \sum_{n=1}^N \mathbb{E}_{z_n, z_{n-1} | {x}, \tilde{{T}} } \left[ \log \mathbb{P} (z_n | z_{n-1}, {T}_0) \right]$$
Given an estimate $\tilde{{T}}$ of the transition matrix, the E-step of Baum-Welch algorithm for clone HMM derives the empirical operator ${T} \mapsto Q_N({T} | \tilde{{T}} )$ by computing the probabilities $ \mathbb{P} (z_n, z_{n-1} | {x}, \tilde{{T}})$. The M-step considers the maximizer operator $M_N: \Omega \to \Omega$ defined by
$M_N(\tilde{{T}}) = \argmax_{{T} \in \Omega} Q_N({T} | \tilde{{T}} )$. That is, the algorithm repeats until convergence: ${T}^{j+1} = M_N({T}^j), \; j \ge 0$.
\bigskip

In the case of an infinite amount of data, the empirical operator $Q_N$ is equivalent to the population function defined as:
$$ \bar{Q}({T} | \tilde{{T}} ) = \lim \limits_{N \to \infty} \mathbb{E} Q_N({T} | \tilde{{T}}).$$
When Assumption \ref{asu1} holds, the following Theorem \ref{main} assess the existence of $\bar{Q}$. This existence is not guaranteed: the quantity $\mathbb{E}Q_N({T} | \tilde{{T}})$ depends upon $N$ since the samples are not i.i.d.. In addition, we define the population maximizer by the self-consistency condition:
$${T}^* = \bar{M}({T}^*) = \arg \max_{{T} \in \Omega } \bar{Q}({T} | {T}^*),$$
where the operator $\bar{M}$ extends $M_N$ to the population function $\bar{Q}$. ${T}^*$ corresponds to a global maximizer of the population likelihood, which we would like to recover via the Baum-Welch algorithm. However, since the samples are not i.i.d., we cannot directly analyze the asymptotic behavior of the population function $\bar{Q}$. To this end, we consider a sequence of latent clones ${z}=(z_{1-k},\ldots, z_{N+k})$ drawn from the stationary distribution $\pi(.|T_0)$ of the Markov chain. For an integer $k$, we introduce the truncated version of the operator $Q_N$:
$$Q_N^k({T} | \tilde{{T}} ) = \frac{1}{N} \sum_{n=1}^N \mathbb{E}_{z_n, z_{n-1} | x_{n-k}, \ldots, x_{n+k}, \tilde{{T}} } \left[ \log \mathbb{P} (z_n | z_{n-1}, {T}) \right].$$
The latent variables $z_n, z_{n-1}$ are conditioned on a $2k$ window centered around the index. Because of the stationarity of the sequence of latent variables, the expectation $\mathbb{E}Q^k_N({T} | \tilde{{T}})$ does not depend upon the sample size $N$. Consequently, we define the truncated population function $\bar{Q}^k$:
$$\bar{Q}^k({T} | \tilde{{T}} ) = \lim \limits_{N \to \infty} \mathbb{E} Q^k_N({T} | \tilde{{T}}) = \mathbb{E} Q^k_1({T} | \tilde{{T}}) .$$
$\bar{Q}^k$ is properly defined. Theorem \ref{main} proves that under Assumption \ref{asu1} and \ref{asu2} (defined below), $\bar{Q}^k$ uniformly approximates the population function $\bar{Q}$. In addition, we extend the operators $M_N$ and $\bar{M}$ to their respective truncated versions:
$$M_N^k(\tilde{{T}} ) = \argmax_{{T} \in \Omega} Q_N^k ({T} | \tilde{{T}} ) \;\;\;\text{     ;    }\;\;\; \bar{M}^k( \tilde{{T}} ) = \argmax_{{T} \in \Omega} \bar{Q}^k ({T} | \tilde{{T}} ).$$
As the truncation level $k$ increases, the operator $M_N^k$ converges to $M_N$. Similarly, as $N$ increases, this operator converges to $\bar{M}^k$. We control their respective convergences with the two following parameters:
\begin{defn}\label{def1}
    Let $\delta \in (0,1)$. Let $\epsilon(N,k, \delta)$ be the smallest scalar such that:
    $$\sup_{{T} \in \Omega}\;\; \mathbb{P}\left[ \| M_N^k({T}) - M_N({T}) \|_2 \ge \epsilon(N,k,\delta) \right] \le \frac{\delta}{2}.$$
    In addition, let $\phi(N,k, \delta)$ be the smallest scalar such that:
    $$\sup_{{T} \in \Omega} \;\; \mathbb{P}\left[ \| M_N^k({T}) - \bar{M}^k({T}) \|_2 \ge \phi(N,k,\delta) \right] \le \frac{\delta}{2}.$$
\end{defn}
We finally make the following assumption.
\begin{asu}\label{asu2}
    For all $\tilde{{T}} \in \Omega$, the truncated population function  $\bar{Q}^k(. | \tilde{{T}})$ is strictly concave. That is,we can fix $\lambda > 0$ such that:
    \begin{equation}\label{rsc}
    \bar{Q}^k({T}_1 | \tilde{{T}} ) -  \bar{Q}^k({T}_2 |\tilde{{T}} ) - \langle \nabla \bar{Q}^k({T}_2 | \tilde{{T}} ), \ {T}_1 - {T}_2 \rangle\le - \frac{\lambda}{2} \| {T}_1 - {T}_2 \|_2^2, \;\;\;\;\; \forall {T}_1, {T}_2, \tilde{{T}} \in \Omega.
    \end{equation}
    In addition for all ${T} \in \Omega$, the gradient of $\bar{Q}^k( {T} | . )$ is Lipschitz with constant $L$. That is:
    \begin{equation}\label{lip}
    \| \nabla \bar{Q}^k({T} | {T}_1) - \nabla \bar{Q}^k({T} | {T}_2) \|_2 \le L \| {T}_1 - {T}_2 \|_2, \;\;\;\;\; \forall T, {T}_1, {T}_2 \in \Omega.
    \end{equation}
    Finally, it holds $\gamma = L / \lambda < 1$.
\end{asu}

We are now ready to present our main Theorem \ref{main}. It states that the difference between the $j$th iterate of the Baum-Welch algorithm $\hat{{T}}^j$ and the global optimum of the population likelihood ${T}^*$ is the sum of a geometrically decaying term and a residual error -- function of the parameters $N$ and $k$. 
\begin{theorem}\label{main}
    Under Assumption \ref{asu1}, the population function $\bar{Q}$ is well-defined.
    
    In addition, if Assumption \ref{asu2} holds, $\bar{M}^k$ converges uniformly to $\bar{M}$. We can fix a sequence $\psi$ such that:
    $$\| \bar{M}^k - \bar{M} \|_{\infty} \le \psi(k) \tolim_{k \to \infty} 0.$$
    Finally, the sequence $\{ \hat{{T}}^j \}_{j\ge 0}$ derived by the Baum-Welch algorithm: $\hat{{T}}^{0} \in \Omega$,  $\hat{{T}}^{j+1} = M_N(\hat{{T}}^j)$, satisfies with probability at least $1 - \delta$:
    \begin{equation}\label{bound-chmm}
    \| \hat{{T}}^j - {T}^*\|_2 \le \gamma^j \| \hat{{T}}^0 - {T}^* \|_2 + \frac{\epsilon(N,k, \delta) + \phi(N,k, \delta) + \psi(k)}{1-\gamma}, \;\;\ j\ge0. 
    \end{equation}
    We define $r(N,k,\delta) = \epsilon(N,k, \delta) + \phi(N,k, \delta) + \psi(k)$ to recover the bound in Theorem \ref{main-shrunk}.
\end{theorem}
The proof is presented in Appendix \ref{sec:proof-main}.

\subsection{Comparing the upper bounds for CHMM and HMM}
We derive herein an upper bound for HMM similar to the one for CHMM in Equation \eqref{bound-chmm} and compare the two upper bounds. Let $E_0$ be the sparse emission matrix associated with a CHMM -- defined such that each clone only emits its observation.  The normalized log-likelihood of the complete data is now lower-bounded by the empirical operator $Q_N^{\textsc{HMM}}$ defined for any couple of estimates $\tilde{{T}}, \tilde{{E}}$ by:
$$Q_N^{\textsc{HMM}}({T}_0, E_0 | \tilde{{T}}, \tilde{{E}} ) = \frac{1}{N} \sum_{n=1}^N \mathbb{E}_{z_n, z_{n-1} | {x}, \tilde{{T}}, \tilde{{E}} } \left[ \log \mathbb{P} (z_n | z_{n-1}, {T}_0, E_0) \right] + \frac{1}{N} \sum_{n=1}^N \mathbb{E}_{z_n | {x}, \tilde{{T}}, \tilde{{E}} } \left[ \log \mathbb{P} (x_n | z_{n}, {T}_0, E_0) \right].$$
By noting that the first term depends only depends upon $\tilde{{T}}$ and the second term only upon $\tilde{{E}}$, we write:
\begin{equation}\label{decomposition}
Q_N^{\textsc{HMM}}({T}, E | \tilde{{T}}, \tilde{{E}} ) = Q_{N,1}^{\textsc{HMM}}({T} | \tilde{{T}}, \tilde{{E}} ) + Q_{N,2}^{\textsc{HMM}}( E | \tilde{{T}}, \tilde{{E}} ).
\end{equation}
We note $\Gamma$ the set of emission matrices. The EM algorithm considers the maximizer operator $M_N: \Omega \times \Gamma \to \Omega \times \Gamma$ and repeats until convergence: $$\left({T}^{j+1}, \; {E}^{j+1} \right) = M^{\textsc{HMM}}_N({T}^{j}, {E}^{j})= \left( M^{\textsc{HMM}}_{N,1}({T}^{j}, {E}^{j}), \; M^{\textsc{HMM}}_{N,2} ({T}^{j}, {E}^{j}) \right),$$
where $M^{\textsc{HMM}}_{N,1}$ and $M^{\textsc{HMM}}_{N,2}$ denote the maximizer operators respectively associated with $Q_{N,1}^{\textsc{HMM}}$ and $Q_{N,2}^{\textsc{HMM}}$.
Similarly to the CHMM case, we define the population versions $\bar{Q}^{\textsc{HMM}}$, $\bar{M}^{\textsc{HMM}}$ of the operators introduced, their truncated versions $Q_N^{\textsc{HMM}, k}$, $M_N^{\textsc{HMM}, k}$ and the truncated population versions $\bar{Q}^{\textsc{HMM}, k}$, $\bar{M}^{\textsc{HMM}, k}$. In particular, the population versions are only defined on a set $\tilde{\Gamma} \subset \Gamma$ of emission matrices with lowest entry lower bounded by some $\beta > 0$. This does not hold for their truncated versions as the truncated operators do not depend upon $N$. In addition, we introduce the operators corresponding to the decomposition with respect to the transition and emission matrices as in Equation \eqref{decomposition}. We consider a global population maximizer $T^*_{\textsc{HMM}}, E^*_{\textsc{HMM}} = \bar{M}^{\textsc{HMM}}(T^*_{\textsc{HMM}}, E^*_{\textsc{HMM}})$ and assume that $T^*_{\textsc{HMM}} = T^*_{\textsc{CHMM}} = T^*$ as defined in Section \ref{sec:ub-chmm}. We finally note $\Theta = \Omega \times \Gamma$, $\tilde{\Theta} = \Omega \times \tilde{\Gamma}$ and extend Assumption \ref{asu2} as follows:

\begin{asu}\label{asu3}
	For all ${\theta} \in \Theta$, the truncated population function  $\bar{Q}^{\textsc{HMM}, k}(. | \theta)$ is strictly concave with respect to each of its component. That is,we can fix $\lambda_1, \lambda_2 > 0$ such that:
	\begin{align}\label{lambda1}
	\begin{split}
	&\bar{Q}^{\textsc{HMM}, k}_1({T}_1 | \theta ) -  \bar{Q}^{\textsc{HMM}, k}_1({T}_2 |{\theta} ) - \langle \nabla \bar{Q}^{\textsc{HMM}, k}({T}_2 | {\theta} ), \ {T}_1 - {T}_2 \rangle\le - \frac{\lambda_1}{2} \| {T}_1 - {T}_2 \|_2^2, \;\;\; \forall {T}_1, {T}_2 \in \Omega, \\
	&\bar{Q}^{\textsc{HMM}, k}_2({E}_1 | {\theta} ) -  \bar{Q}^{\textsc{HMM}, k}_2({E}_2 |{\theta} ) - \langle \nabla \bar{Q}^{\textsc{HMM}, k}({E}_2 | {\theta} ), \ {E}_1 - {E}_2 \rangle\le - \frac{\lambda_2}{2} \| {E}_1 - {E}_2 \|_2^2, \;\;\; \forall {E}_1, {E}_2 \in \Gamma.
	\end{split}
	\end{align}
	In addition for all ${T} \in \Omega$, the gradient of $\bar{Q}_1^{\textsc{HMM}, k}( {T} | . )$ is Lipschitz with respect to each variable:
	\begin{align}\label{L11}
	\begin{split}
	&\| \nabla \bar{Q}_1^{\textsc{HMM}, k}({T} | {T}_1, E_1) - \nabla \bar{Q}_1^{\textsc{HMM}, k}({T} | {T}_2, E_1) \|_2 \le L_{1,1} \| {T}_1 - {T}_2 \|_2, \;\;\; \forall {T}_1, {T}_2 \in \Omega, \;\;\; \forall {E}_1 \in \Gamma,\\
	&\| \nabla \bar{Q}_1^{\textsc{HMM}, k}({T} | {T}_1, E_1) - \nabla \bar{Q}_1^{\textsc{HMM}, k}({T} | {T}_1, E_2) \|_2 \le L_{2,1} \| {E}_1 - {E}_2 \|_2, \;\;\; \forall {T}_1 \in \Omega, \;\;\; \forall {E}_1, {E}_2 \in \Gamma.
	\end{split}
	\end{align}
	We define two similar conditions for $\nabla \bar{Q}_2^{\textsc{HMM}, k}( E | . )$ -- with respective Lipschitz constants $L_{1,2}$ and $L_{2,2}$.
	
	We note $L^{\textsc{HMM}} = \max(L_{1,1}, L_{2,1}) + \max(L_{1,2}, L_{2,2})$ and $\lambda^{\textsc{HMM}} = \min(\lambda_1, \lambda_2)$ and assume: $$\gamma_{\textsc{HMM}} = L^{\textsc{HMM}} / \lambda^{\textsc{HMM}} < 1.$$
\end{asu}
We finally extend Definition \ref{def1} as follows:
\begin{defn}
	Let $\delta \in (0,1)$. Let $\epsilon^{\textsc{HMM}}(N,k, \delta)$ be the smallest scalar such that:
	$$\sup_{{\theta} \in \Theta}\;\; \mathbb{P}\left[ \| M_N^{\textsc{HMM},k}({\theta}) - M_N^{\textsc{HMM}}({\theta}) \|_2 \ge \epsilon^{\textsc{HMM}}(N,k,\delta) \right] \le \frac{\delta}{2}.$$
	In addition, let $\phi^{\textsc{HMM}}(N,k, \delta)$ be the smallest scalar such that:
	$$\sup_{{\theta} \in \Theta} \;\; \mathbb{P}\left[ \| M_N^{\textsc{HMM},k}({\theta}) - \bar{M}^{\textsc{HMM},k}({\theta}) \|_2 \ge \phi^{\textsc{HMM}}(N,k,\delta) \right] \le \frac{\delta}{2}.$$
\end{defn}

We derive the following theorem in the case of HMM. The proof is presented in Appendix \ref{sec:proof-main}.
\begin{theorem}\label{main-hmm}
	Under Assumption \ref{asu1}, the population function $\bar{Q}^{\textsc{HMM}}$ is well-defined on $\tilde{\Theta} \times \tilde{\Theta}$. In addition, if Assumption \ref{asu3} holds, $\bar{M}^{\textsc{HMM}, k}$ converges uniformly to $\bar{M}^{\textsc{HMM}}$: $\| \bar{M}^{\textsc{HMM}, k} - \bar{M}^{\textsc{HMM}} \|_{\infty} \le \psi^{\textsc{HMM}}(k) \tolim \limits_{k \to \infty} 0.$
	
	Finally, the sequence of transition matrices $\{ \hat{{T}}_{\textsc{HMM}}^j \}_{j\ge 0}$ derived by the Baum-Welch algorithm: $\hat{{T}}^{0} \in \Omega$,  $\hat{{T}}_{\textsc{HMM}}^{j+1} = M_{N,1}(\hat{{T}}^j, \hat{{E}}^j)$, satisfies with probability at least $1 - \delta$:
	\begin{equation}\label{bound-hmm}
	\| \hat{{T}}_{\textsc{HMM}}^j - {T}^*\|_2 \le \gamma_{\textsc{HMM}}^j \| \hat{{T}}^0 - {T}^* \|_2 + \frac{\epsilon^{\textsc{HMM}}(N,k, \delta) + \phi^{\textsc{HMM}}(N,k, \delta) + \psi^{\textsc{HMM}}(k)}{1-\gamma_{\textsc{HMM}}}, \;\;\ j\ge0. 
	\end{equation}
	We define $r_{\textsc{HMM}}(N,k,\delta) = \epsilon^{\textsc{HMM}}(N,k, \delta) + \phi^{\textsc{HMM}}(N,k, \delta) + \psi^{\textsc{HMM}}(k)$ to recover the bound in Theorem \ref{comparison}. It holds:
	$$ \gamma_{\textsc{CHMM}}  < \gamma_{\textsc{HMM}}  \text{ and } r_{\textsc{CHMM}}(N,k,\delta) < r_{\textsc{HMM}}(N,k,\delta) .$$
\end{theorem}

\subsection{Proof of Theorem \ref{main}}\label{sec:proof-main}
\begin{Proof}
    We divide our proof in 3 steps.

    \textbf{Step 1: } We first prove the existence of the population function $\bar{Q}$ defined as: $\bar{Q}(. | . ) = \lim \limits_{N \to \infty} \mathbb{E} Q_N(. | . ).$
    To do so, we prove that the sequence of operators $\{\mathbb{E} Q_N\}_{N\ge0}$ satisfies the Cauchy property:
    $$\forall \epsilon, \;\; \exists R\ge 0: \;\; \forall N, M \ge R, \;\; \|\mathbb{E} Q_N - \mathbb{E} Q_M \|_{\infty} \le \epsilon.$$
    Following Lemma 3 from \cite{yang2017statistical}, since the Markov chain ${z}$ with stationary distribution $\pi(.|T_0)$ satisfies Assumption \ref{asu1}, it holds for some $C_0 > 0$:
    \begin{equation}\label{eq-sup}
    \sup_{{T} \in \Omega} \sup_{{x}} \sum_{z_n} \lvert \mathbb{P}(z_n | x_1, \ldots, x_N, {T}) - \mathbb{P}(z_n |x_{n-k}, \ldots, x_{n+k}, {T}) \rvert \le C_0 (1 - \epsilon \pi_{\min})^k, \;\; \forall n \in \left[1, N \right]
    \end{equation}
    where $\pi_{\min} = \min_{j \le K, {T} \in \Omega} \pi(j| {T})$. 
    Consequently, since $\bar{Q}^k(.|.) = \mathbb{E} Q_N^k(.|.)$, it holds $\forall N, \; \forall {T}, \tilde{{T}} \in \Omega$:
    \begin{align}
    \begin{split}
    &\lvert \mathbb{E} Q_N({T} | \tilde{{T}} ) - \bar{Q}^k({T} | \tilde{{T}} ) \rvert\\
    &= \frac{1}{N} \bigg\lvert \sum_{n=1}^N \sum_{z_n, z_{n-1}} \mathbb{E} \left[ \left\{ \mathbb{P}(z_n, z_{n-1} | x_1, \ldots, x_N, \tilde{{T}}) - \mathbb{P}(z_n, z_{n-1} | x_{n-k}, \ldots, x_{n+k}, \tilde{{T}}) \right\} \log \mathbb{P} (z_n | z_{n-1}, {T}) \right]\bigg\rvert  \\
    &\le \frac{1}{N} \sum_{n=1}^N \sum_{z_n, z_{n-1}}  \mathbb{E} \big\rvert \left\{ \mathbb{P}(z_{n-1} | x_1, \ldots, x_N, \tilde{{T}}) - \mathbb{P}(z_{n-1} | x_{n-k}, \ldots, x_{n+k}, \tilde{{T}}) \right\} \mathbb{P}(z_n | z_{n-1},\tilde{{T}}) \log \mathbb{P} (z_n | z_{n-1}, {T}) \big\lvert \\
    &\le \frac{1}{N} \sum_{n=1}^N \sum_{z_{n-1}} \sup_{\tilde{{T}}, {x}} \big\rvert  \mathbb{P}(z_{n-1} | x_1, \ldots, x_N, \tilde{{T}}) - \mathbb{P}(z_{n-1} | x_{n-k}, \ldots, x_{n+k}, \tilde{{T}}) \big\lvert \; \mathbb{E} \bigg\rvert \sum_{z_n}\mathbb{P}(z_n | z_{n-1},\tilde{{T}}) \log \mathbb{P} (z_n | z_{n-1}, {T}) \bigg\lvert \\
    &\le  C_0 (1 - \epsilon \pi_{\min})^k  \frac{1}{N} \sum_{n=1}^N \max_{z_{n-1}} \mathbb{E} \bigg\rvert \sum_{z_n}\mathbb{P}(z_n | z_{n-1},\tilde{{T}}) \log \mathbb{P} (z_n | z_{n-1}, {T}) \bigg\lvert\\
    &\le  C_0 K \log(\pi_{\min}^{-1})  (1 - \epsilon \pi_{\min})^k \\
    &\le  C_1  (1 - \epsilon \pi_{\min})^k \text{ for } C_1 = C_0 K \log(\pi_{\min}^{-1}).
    \end{split}
    \end{align}
    As a result we have:
    \begin{equation}\label{bound}
    \|\mathbb{E} Q_N - \bar{Q}^k \|_{\infty} \le C_1  (1 - \epsilon \pi_{\min})^k.
    \end{equation}
    Using the triangle inequality: 
    $$\|\mathbb{E} Q_N - \mathbb{E} Q_{M} \|_{\infty} \le \|\mathbb{E} Q_M - \bar{Q}^k \|_{\infty} + \| \bar{Q}^k - \mathbb{E} Q_{M} \|_{\infty} \le 2C_1  (1 - \epsilon \pi_{\min})^k,$$ 
    we conclude that if $k \ge \left\lceil \frac{\log (\epsilon / 2 C_1)}{\log(1- \epsilon \pi_{\min})} \right\rceil$, the sequence $\{\mathbb{E} Q_N\}_{N\ge0}$ is Cauchy and $\bar{Q}$ is well-defined.
    
    \paragraph{Step 2: } We now prove the uniform convergence of the truncated operator $\bar{M}^k$ to its population version $\bar{M}$. Setting $N \to \infty$ in Equation \eqref{bound} gives:
    $$\bar{Q}^k(\tilde{{T}} | {T}) \le \bar{Q}(\tilde{{T}} | {T}) + C_1  (1 - \epsilon \pi_{\min})^k \;\;  \forall {T}, \tilde{{T}} \in \Omega.$$
    Let ${T} \in \Omega$. Since $\bar{M}( {T} ) = \argmax_{\tilde{{T}} \in \Omega} \bar{Q} (\tilde{{T}} | {T} )$ it holds:
    $$\bar{Q} (\bar{M}( {T} )| {T} )\ge \bar{Q} (\bar{M}^k( {T} )| {T} )\ge \bar{Q}^k (\bar{M}^k( {T} )| {T} ) - C_1  (1 - \epsilon \pi_{\min})^k. $$
    In addition, Equation \eqref{bound} also gives:
    $$ C_1  (1 - \epsilon \pi_{\min})^k \ge \bar{Q} (\bar{M}( {T} )| {T} ) - \bar{Q}^k (\bar{M}( {T} )| {T} ).$$
    Hence, by combining the previous relations and using the strict concavity of $\bar{Q}^k (.| {T} )$ it holds:
    \begin{align}\label{almost-done}
    \begin{split}
    2C_1 (1 - \epsilon \pi_{\min})^k 
    &\ge \bar{Q}^k (\bar{M}^k( {T} )| {T} ) - \bar{Q}^k (\bar{M}( {T} )| {T} )   \\
    & \ge \frac{\lambda}{2} \| \bar{M}^k( {T} ) - \bar{M}( {T} )\|_2^2 + \langle \nabla \bar{Q}^k( \bar{M}^k( {T} )  | {T} ), \  \bar{M}^k( {T} ) - \bar{M}( {T} )\rangle.
    \end{split}
    \end{align}
    Finally, since $\bar{M}^k( {T} ) = \argmax_{\tilde{{T}} \in \Omega} \bar{Q}^k (\tilde{{T}} | {T} )$, the first order condition for optimality gives:
    \begin{equation}\label{eq0}
    \langle \nabla \bar{Q}^k( \bar{M}^k( {T} )  | {T} ), \  \bar{M}^k( {T}) - \tilde{{T}} ) \rangle \ge 0, \;\; \forall \tilde{{T}} \in \Omega.
    \end{equation}
    We fix $\tilde{{T}} = \bar{M}( {T} )$ and inject in Equation \eqref{almost-done} to conclude:
    \begin{equation}\label{conclusion}
    \| \bar{M}^k( {T} ) - \bar{M}( {T} )\|_2^2
    \le \psi(k)^2 := \frac{4 C_1}{\lambda} (1 - \epsilon \pi_{\min})^k \;\;\text{ for all } {T}\in \Omega,
    \end{equation}
    which proves the universal convergence of $\bar{M}^k$ to $ \bar{M}$ -- all norms are equivalent in the finite space $\Omega$.

    \paragraph{Step 3: } Our final steps derive a control for the difference between the $j$th Baum-Welch estimate of the transition matrix $\hat{{T}}^{j}$ and the ground-truth one ${T}^*$.
    \bigskip
    
    We first show that the application ${T} \mapsto \bar{M}^k( {T} )$ is contractive. Let us fix ${T}, \tilde{{T}} \in \Omega$. Using the first order condition for optimality in Equation \eqref{eq0} for $\tilde{{T}} = \bar{M}^k( \tilde{{T}} )$, it holds: 
    \begin{equation}\label{eq1}
    \langle \nabla \bar{Q}^k( \bar{M}^k( {T} )  | {T} ), \ \bar{M}^k( {T} ) - \bar{M}^k( \tilde{{T}} ) \rangle \ge 0
    \end{equation}
    Similarly, we have $\bar{M}^k( {T} ) = \argmax_{\tilde{{T}} \in \Omega} \bar{Q}^k ( \tilde{{T}} | {T} )$ thus by switching the role of $
    \bar{M}^k( {T} )$ and $\bar{M}^k( \tilde{{T}} )$:
    \begin{equation}\label{eq2}
    \langle \nabla \bar{Q}^k( \bar{M}^k( \tilde{{T}} )  | \tilde{{T}} ), \  \bar{M}^k( \tilde{{T}} ) - \bar{M}^k( {T} )  \rangle \ge 0, \;\; \forall {T} \in \Omega.
    \end{equation}
    Summing Equations~\eqref{eq1} and~\eqref{eq2} we have:
    \begin{equation}\label{eq3}
    \langle \nabla \bar{Q}^k( \bar{M}^k( \tilde{{T}} )  | \tilde{{T}} ) - \nabla \bar{Q}^k( \bar{M}^k( {T} )  | {T} ),  \ \bar{M}^k( \tilde{{T}} )-  \bar{M}^k( {T} ) \rangle \ge 0
    \end{equation}
    In addition, Assumption \ref{asu2} guarantees that $\bar{Q}^k(. | \tilde{{T}})$ is strictly concave. By switching the role of ${T}_1$ and ${T}_2$ in Equation \eqref{rsc} and summing the two inequalities it holds:
    $$\langle \nabla \bar{Q}^k({T}_1 | \tilde{{T}} ), \ {T}_1 - {T}_2 \rangle - \langle \nabla \bar{Q}^k({T}_2 | \tilde{{T}} ), \ {T}_1 - {T}_2 \rangle\le - \lambda \| {T}_1 - {T}_2 \|_2^2, \;\;\; \forall {T}_1, {T}_2 \in \Omega,$$
    which in the case where ${T}_1= \bar{M}^k( {T} ) $ and  ${T}_2= \bar{M}^k( \tilde{{T}} )$ is equivalent to saying that:
    \begin{equation}\label{eq4}
    \lambda \| \bar{M}^k( {T} )  - \bar{M}^k( \tilde{{T}} )  \|_2^2 \le
    \langle \nabla \bar{Q}^k(\bar{M}^k( {T} )  | \tilde{{T}} ) - \nabla \bar{Q}^k(\bar{M}^k( \tilde{{T}} )  | \tilde{{T}} ), \ \bar{M}^k( \tilde{{T}} )  - \bar{M}^k( {T} ) \rangle .
    \end{equation}
    Therefore, by summing Equations \eqref{eq3} and  \eqref{eq4} and by using the Lipschitz condition, it holds:
    \begin{align*}
    \lambda \| \bar{M}^k( {T} )  - \bar{M}^k( \tilde{{T}} )  \|_2^2
    &\le \langle \nabla \bar{Q}^k(\bar{M}^k( {T} )  | \tilde{{T}} ) - \nabla \bar{Q}^k(\bar{M}^k( {T} )  | {T} ), \ \bar{M}^k( \tilde{{T}} )  - \bar{M}^k( {T} ) \rangle \\
    & \le \| \nabla \bar{Q}^k(\bar{M}^k( {T} )  | \tilde{{T}} ) - \nabla \bar{Q}^k(\bar{M}^k( {T} )  | {T} ) \|_2 \| \bar{M}^k( {T} )  - \bar{M}^k( \tilde{{T}} )  \|_2 \\
    &\;\;\;\;\;\; \text{ with Cauchy-Schwartz inequality}\\
    & \le  L \|   {T} - \tilde{{T}} \|_2  \| \bar{M}^k( {T} )  - \bar{M}^k( \tilde{{T}} )  \|_2,
    \end{align*}
    which is equivalent to saying that:
     \begin{equation}\label{eq11}
         \| \bar{M}^k( {T} )  - \bar{M}^k( \tilde{{T}} )  \|_2 \le \gamma  \|  {T} - \tilde{{T}} \|_2,
     \end{equation}
    where $\gamma = L / \lambda <1$. Hence the application ${T} \mapsto \bar{M}^k( {T} )$ is contractive.

    \bigskip
    Second, let us note that the population maximizer ${T}^*$ is a fixed point for the operator $\bar{M}$ but not for $\bar{M}^k$ -- because of the truncation level.  In addition, in the particular case where ${T}={T}^*$, and by using the self-consistency condition $\bar{M}( {T}^*) ={T}^*$, Equation \eqref{conclusion} implies:
    \begin{equation}\label{eq12}
    \| \bar{M}^k( {T}^* ) - {T}^* \|_2 \le \psi(k).
    \end{equation}
    Consequently, by combining Equations \eqref{eq11} and \eqref{eq12}, the sequence $\{ \tilde{{T}}^j \}_{j\ge 0}$ generated by the $\bar{M}^k$ operator: $\tilde{{T}}^{j+1} = \bar{M}^k(\tilde{{T}}^{j})$ satisfies:
    \begin{align*}
    \| \tilde{{T}}^{j} - {T}^* \|_2
    &= \| \bar{M}^k(\tilde{{T}}^{j-1})  - {T}^* \|_2 \\
    &\le \| \bar{M}^k(\tilde{{T}}^{j-1})  - \bar{M}^k({T}^*) \|_2 + \| \bar{M}^k({T}^*) - {T}^* \|_2 \\
    &\le \gamma \| \tilde{{T}}^{j-1}  -  {T}^* \|_2 + \psi(k).
    \end{align*}

 Finally, we consider the sequence $\{ \hat{{T}}^j \}_{j\ge 0}$ drawn by the EM operator $M_N$. Combining our previous results, it holds with probability at least $1 - \delta$:
\begin{align}\label{geo}
\begin{split}
\| \hat{{T}}^{j} - {T}^* \|_2
&= \| M_N(\hat{{T}}^{j-1}) - {T}^* \|_2 \\
&\le \| M_N(\hat{{T}}^{j-1}) - M_N^k(\hat{{T}}^{j-1}) \|_2
+ \| M_N^k(\hat{{T}}^{j-1}) - \bar{M}^k(\tilde{{T}}^{j-1}) \|_2
+ \| \bar{M}^k(\tilde{{T}}^{j-1})  - {T}^* \|_2\\
&\le \epsilon(N,k,\delta) + \psi(N,k,\delta) + \gamma \| \tilde{{T}}^{j-1}  -  {T}^* \|_2 + \psi(k).
\end{split}
\end{align}
By summing Equation \eqref{geo} it holds with probability at least $1 - \delta$:
\begin{align*}
\| \hat{{T}}^{j} - {T}^* \|_2
\le \gamma^j \| \tilde{{T}}^{0}  -  {T}^* \|_2 + \sum_{\ell=0}^j \gamma^{\ell} \{ \epsilon(N,k,\delta) + \psi(N,k,\delta) + \psi(k) \}\\
\le \gamma^j \| \tilde{{T}}^{0}  -  {T}^* \|_2 + \frac{1}{1 - \gamma} \{ \epsilon(N,k,\delta) + \psi(N,k,\delta) + \psi(k) \},
 \end{align*}
 which concludes the proof.
 \end{Proof}

\subsection{Proof of Theorem \ref{main}}\label{sec:proof-hmm}
The first part of the theorem and the upper bound follows from Proposition 1, Theorem 1 and Theorem 2 in \cite{yang2017statistical} since the mixing condition, the strong-concavity condition, the Lipchitz condition are satisfied and it holds:
$$\sup_{E \in \tilde{\Gamma}} \mathbb{E} \left[\max_{j \le K} | \log | \mathbb{P}(x_i | z_i, E) |  \right] < \infty.$$

We now assume that both Equations \eqref{bound-chmm} and \eqref{bound-hmm} are satisfied for CHMM and HMM. Equation \eqref{L11} holds for ${E}_1 \in \Gamma$. In the particular case where $E_1 = E_0$ is the CHMM transition matrix, we have $L^{\textsc{CHMM}} \le L_{11}$. Consequently  $L^{\textsc{CHMM}} < L^{\textsc{HMM}}$.

Similarly, since Equation \eqref{lambda1} holds for all $\theta$ of the form $\theta = (T, E_0)$ then it holds $\lambda^{\textsc{CHMM}} \ge \lambda^{\textsc{HMM}}$.

By pairing both results we obtain: $$\gamma^{\textsc{CHMM}} < \gamma^{\textsc{HMM}}.$$

In addition, since for $T \in \Omega$ we have $ \| M_N^{\textsc{CHMM}, k}({T}) - M_N^{\textsc{CHMM}, k}({T}) \|_2 = \| M_N^{\textsc{HMM}, k}({T, E_0}) - M_N^{\textsc{HMM}, k}({T, E_0}) \|_2$, then  $\epsilon^{\textsc{CHMM}}(N,k, \delta) \le \epsilon^{\textsc{HMM}}(N,k, \delta)$. Similarly $\psi^{\textsc{CHMM}}(N,k, \delta) \le  \psi^{\textsc{HMM}}(N,k, \delta)$.

In addition, we have defined in Section \ref{sec:ub-chmm}. $\psi^{\textsc{CHMM}}(k)= 4 C_0 K \log(\pi_{\min}^{-1}) (1 - \epsilon \pi_{\min})^k / \lambda^{\textsc{CHMM}}$. Since $\psi^{\textsc{HMM}}(k)$ is of the same form \cite{yang2017statistical}, it holds $\psi^{\textsc{HMM}}(k) < \psi^{\textsc{CHMM}}(k)$.

Hence, we conclude the proof by combining the previous relations:
$$r^{\textsc{CHMM}}(N,k,\delta) < r^{\textsc{HMM}}(N,k,\delta).$$

\newpage
\bibliographystyle{plain}
\bibliography{chmm}
\end{document}